\documentclass[journal]{IEEEtran}
\usepackage{subfigure}
\usepackage{graphicx}
\usepackage{amsmath}
\usepackage{amsthm}
\usepackage{booktabs}
\usepackage{algorithm}
\usepackage{algorithmic}
\usepackage{subfigure}
\usepackage{multirow}
\usepackage{url}

\usepackage{bm}

\begin{document}

	\title{PHI-MVS: Plane Hypothesis Inference Multi-view Stereo for Large-Scale Scene Reconstruction }

	\author{
		Shang Sun,Yunan Zheng, Xuelei Shi, Zhenyu Xu, Yiguang Liu*
			
	}

	\maketitle
	
	\begin{abstract}
	PatchMatch based Multi-view Stereo (MVS) algorithms have achieved great success in large-scale scene reconstruction tasks. However, reconstruction of texture-less planes often fails as similarity measurement methods may become ineffective on these regions.  Thus, a new plane hypothesis inference strategy is proposed to handle the above issue. The procedure consists of two steps: First,  multiple plane hypotheses are generated using filtered initial depth maps on regions that are not successfully recovered; Second, depth hypotheses are selected using  Markov Random Field (MRF). The strategy can significantly improve the completeness of reconstruction results with only acceptable computing time increasing. Besides, a new acceleration scheme similar to dilated convolution can speed up the depth map estimating process with only a slight influence on the reconstruction. We integrated the above ideas into a new MVS pipeline, Plane Hypothesis Inference Multi-view Stereo (PHI-MVS). The result of PHI-MVS is validated on ETH3D public benchmarks, and it demonstrates competing performance against the state-of-the-art.
	\end{abstract}
	

	\IEEEpeerreviewmaketitle
	
	\section{Introduction}
	\IEEEPARstart{M}{ulti-view stereo(MVS)} is an important research field in computer vision, and it aims to recover the three-dimensional structure of scenes using 2D images and camera parameters. The camera parameters (both internal and external parameters)  can be obtained from the Structure From Motion software (such as Visual SFM\cite{wu2013towards}, COLMAP\cite{schonberger2016structure}) or manual calibration. Nowadays, MVS can be used in many applications \cite{2007MonoSLAM,2011DTAM,2013Optical,2013Learning} related to computer vision.  Thanks to the public datasets and their online evaluation systems, it is becoming easier for authors to evaluate their algorithms. Depth-map merging based MVS aims to recover depth value for each pixel on every image and then merge them to obtain a final point cloud. The most popular way to implement stereo matching in the depth-map merging based MVS algorithm nowadays should be the PatchMatch\cite{barnes2009patchmatch}. \par 
	
	However, when encountering texture-less regions, PatchMatch based method may become ineffective. There are two main reasons for the failure: First, to perform efficient stereo matching, local similarity measurement methods such as Normalized Cross Correlation(NCC) are widely used in the stereo matching step.  As these methods can only reflect local similarity, there may exist plenty of wrong depth hypotheses whose similarity scores are even higher than that of ground truth on texture-less regions. Second,  during the whole pipeline, random refinement is necessary to help the result jump out of local minimal. However, this procedure may also cause the result of texture-less regions to jump into wrong depth values whose similarity scores are higher than that of ground truth. Thus, we use an additional module to efficiently generate plane hypotheses with the filtered depth values output by stereo matching step, then choosing which hypothesis should be preserved by MRF. As the label number is small, the MAP inference of MRF can be executed within an acceptable time. Later, geometric consistency based filtering procedure is applied to remove outliers.  \par
	We know that the operation of accessing memory in the GPU unit consumes much time.  Furthermore, by increasing support window size during stereo matching, the pixels' numbers will increase rapidly. To speed up our algorithm, we borrowed the idea from dilated convolution \cite{yu2015multi} during the stereo matching step. As shown in the experiment, our speed-up scheme is effective and can boost the algorithm with only a slight influence on the result.\par

		\begin{figure}
		\setlength{\belowcaptionskip}{-0.3cm} 
		\centering
		\subfigure[]{
			\includegraphics[scale=.39]{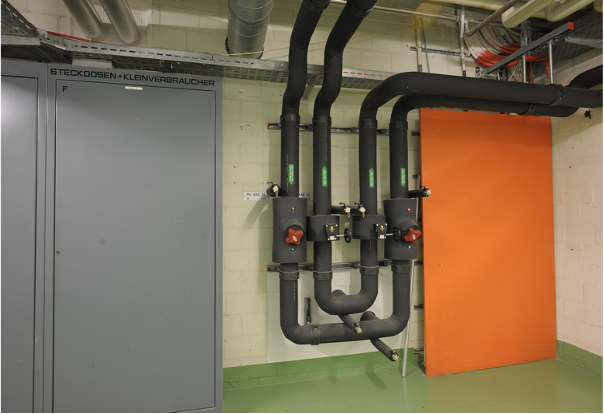}
		}
		\subfigure[]{
			\includegraphics[scale=.39]{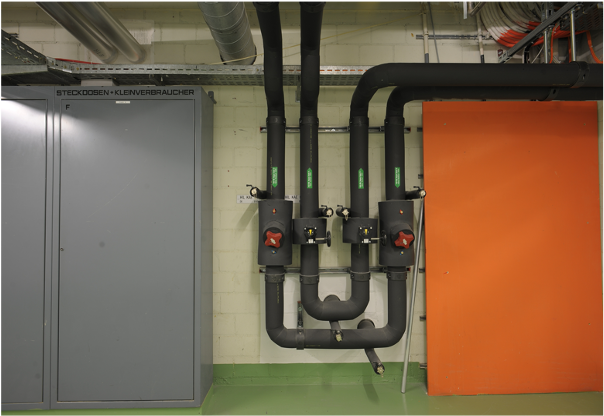}
		}
		\subfigure[]{
			\includegraphics[scale=.39]{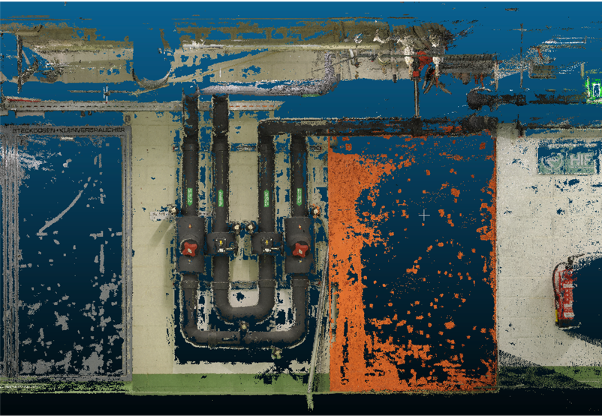}
		}
		\subfigure[]{
			\includegraphics[scale=.39]{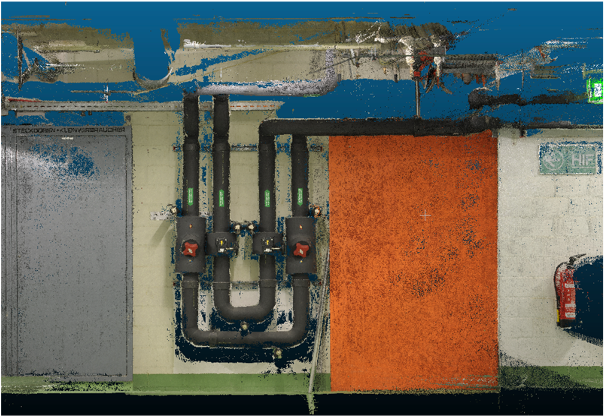}
		}
		
		\caption{(a) and (b) are input images of ETH3D dataset. (c) is the point cloud result without the plane hypothesis inference procedure, (d) is the result of our method. }
		\label{FIG:pipeResult}
	\end{figure}

	Thus, we present PHI-MVS (Plane Hypothesis Inference Multi-View Stereo), a new MVS pipeline that has integrated our above ideas.  The main contributions of PHI-MVS are as follows:
	\begin{itemize}
		\item A plane hypothesis  generating  and selecting strategy using MRF, which can improve the reconstruction on texture-less regions. The additional module can greatly increase the completeness of reconstruction with acceptable computing time consumption.     
		\item A scheme similar to dilated convolution can speed up the stereo matching step with only a slight influence on the accuracy and the completeness of results.
		\item A concise and practical MVS pipeline is proposed and the results on  ETH3D high-resolution data set show its state-of-the-art performance. 
	\end{itemize}

	\section{Related Work}
	MVS algorithms can be divided into four categories according to \cite{seitz2006comparison}, they are voxel-based methods \cite{2006Multi,2014Probabilistic}, surface evolution-based methods \cite{2003Silhouette}, feature point growing-based methods \cite{2010Accurate,Engin2011Efficient}, and depth-map merging based methods \cite{2007Multi,2017fast}. The proposed method belongs to the fourth category.   Depth map-based MVS methods obtain a 3D dense model in two stages. First, the depth map is estimated for each reference image using stereo matching algorithms. Second, outliers are filtered, and a 3D dense model is obtained by merging all back-projected pixels. \par
	In the early years, many algorithms \cite{1999Fast,2004An,kolmogorov2006convergent,campbell2008using} use the Markov Random Field to solve stereo problems. These algorithms' main steps are as follows: First, they compute matching costs with discrete depth values or disparities for each pixel. Second, an energy equation that can be used to perform MAP inference is built. The energy function usually contains unary items and pairwise items. Unary items usually correspond to matching similarity, and pairwise items usually correspond to depth values' continuity and smoothness. Third, minimize the energy equation.  As in general cases, minimize the energy equation is an NP-hard problem. Approximate minimization techniques such as \cite{2004What,2001Generalized,kolmogorov2006convergent}  are usually used to solve the problem. The MRF-based methods have achieved satisfactory results as it is obvious that the depth values should be smooth and continuous between most neighboring pixels, except in the edge part. However, solving the energy function consumes a lot of memory and time. With the increase of image resolution and depth accuracy requirements, it is impossible to perform MRF-based methods within acceptable memory and time limits.\par
	PatchMatch based algorithms gradually replaced the MRF-based algorithms in recent years. The first one that uses the PatchMatch algorithm to solve the stereo problem is \cite{bleyer2011patchmatch}. Then \cite{shen2013accurate} introduces this method to MVS problems. \cite{galliani2015massively} is the first to implement the PatchMatch based algorithm on GPUs, and it uses the black-red pattern in the spatial spreading step. Nowadays, many  algorithms\cite{xu2020marmvs,li2020confidence,xu2020planar} have achieved satisfactory performance. The main part of the whole pipeline is as follows: First, initializing the depth and normal values for each pixel using the random technique. Second, selecting neighboring images for each reference image. Third, performing stereo matching and spread good depth or normal hypothesis to neighboring pixels. Fourth, refining the depth and normal values with random disturb. Fifth, filtering the depth and normal vectors via geometric consistency. Last, back projecting pixels and merging them to a point cloud model. In each step, it may vary from one to another.\par
	Deep learning has a significant influence on 3D vision. Recently, methods based on deep learning \cite{yao2018mvsnet,yi2020pyramid,xue2019mvscrf,luo2020attention,2020Mesh,li2020confidence} have achieved inspiring results. They can be roughly divided into two categories. The first category is the end-to-end methods as \cite{yao2018mvsnet,luo2020attention}. The main pipeline of these methods is as follows: First,  use convolution networks to extract the original images' features. Transform pixels on the reference image to corresponding pixels on source images with discrete depth or disparity hypothesis. Compute the similarity of each depth hypothesis using the extracted features and obtain a cost volume. Second,  different style neural networks are used to estimate the depth of the reference image based on the cost volume. Though they perform well on data sets \cite{aanaes2016large,Knapitsch2017}, many end-to-end methods could not work on ETH high-resolution benchmark according to \cite{xu2020pvsnet}. Here are the reasons: Cost volume needs finite discrete depth values. With the increase of image resolution and depth accuracy requirements, these algorithms have to downsample the original images because of limited GPU memory. Besides, large viewpoint change breaks the assumption that neighboring images have a strong visibility association with the reference image. Though \cite{xu2020pvsnet} try to handle the above issues, their performance on ETH3D still slightly lagging behind traditional methods. Other kinds of methods as \cite{2020Mesh,li2020confidence} use CNN as additional modules in the whole pipeline. Most of these methods aim to use CNN to predict the confidence of the depth and normal maps and then filter or pad the results using the confidence. \par 
	It is hard for PatchMatch based methods to recover correct depth information on texture-less regions. Many previous works\cite{romanoni2019tapa,2020Mesh,li2020confidence,2020Mesh} have aimed to improve the reconstruction of these regions. However, most of these works either need segmentation information of original color images or based on data-driven deep learning methods. \cite{romanoni2019tapa} handles this issue mainly by subdividing the image into superpixels. \cite{li2020confidence} estimates the confidence of the reconstructed depth map and then applies an interpolation stage. \cite{2020Mesh} uses the projection of points recovered by mesh and then uses CNN to filter outliers. \cite{xu2020planar} uses planar prior as guidance and geometric consistency during stereo matching.\par

	\section{ PHI-MVS: the method}
	Our algorithm can mainly be divided into six parts:  neighboring image selection, random initialization,  depth map estimation,  filtering, plane hypothesis inference, and fusion. 
	We implement our method on GPUs with CUDA, and adopt the red-black pattern updating scheme in the spatial spreading step. To further speed up the algorithm, we adopt the improved red-black sampling method proposed by \cite{xu2019multi}. The goal of our whole pipeline is to estimate depth and normal values$(d_{\bm{x}},\bm{n_x})$ for each pixels $\bm{x}$ on each reference image and then merge them to a single point cloud.  The whole pipeline of the PHI-MVS is briefly shown in Algorithm \ref{pipeline}.
	
	\subsection{Neighboring Image Selection}
	
	Choosing neighboring images for each reference image is an important part of the whole pipeline.  Source images that are close to the reference image may have more overlapping regions, and using them can increase the completeness of the reconstructed result,  but it may introduce more errors as large depth change can only cause slight pixel position displacement on source images. Source images that are far from the reference image may improve the reconstruction's accuracy, but they may have fewer overlapping regions and more occlusions, which may decrease the result's completeness. \cite{shen2013accurate} selects neighboring images using the length of baseline and the angle between principal axes of cameras. It computes the  angle $\vartheta_{ij}$ between principal axes and the baseline length   $b_{ij}$ between the reference image $i$ and the source image  $j$, and filters source images whose  $5^{\circ}  \leq \vartheta_{ij}   \leq 60^{\circ} $   and $ 0.05\bar{b}  \leq b_{ij}  \leq  2\bar{b}$, $\bar{b} $ is the median baseline length  of the reference image $i$ and all of other source images. It sorts source images by $\vartheta_{ij}b_{ij}$ from small to large. However, its super parameters are hard to determine when the scale of scenes and the number of total images varies. The parameter used in \cite{shen2013accurate} even could not work for some scenes in \cite{schoeps2017cvpr}.   The strategy used in \cite{xu2020marmvs}  first filters source images using  $\tau $.   $\tau$ is the projection point's position displacement when given 3D points a disturb. Its implementation  is  shown as  Eq. \ref{EQ:disturb} and Eq. \ref{EQ:filetrWithXu}.

	\begin{figure}
		\setlength{\belowcaptionskip}{-0.4cm}
		\setlength{\abovecaptionskip}{-0.05cm} 
		\centering
		\subfigure[]{
			\includegraphics[scale=.17]{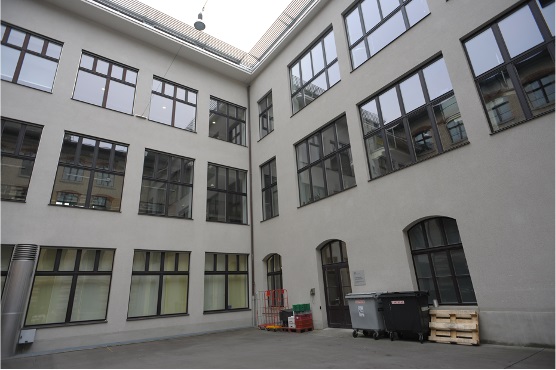}	
		}
		\subfigure[]{
			\includegraphics[scale=.17]{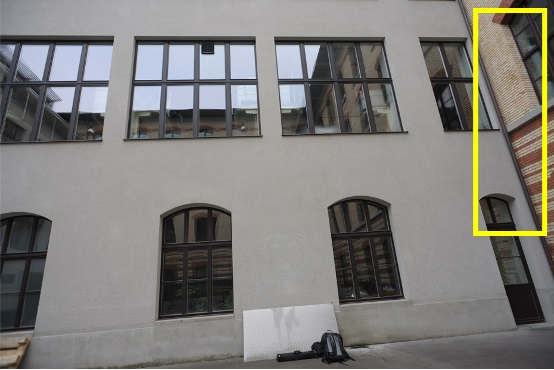}	
		}	\subfigure[]{
			\includegraphics[scale=.17]{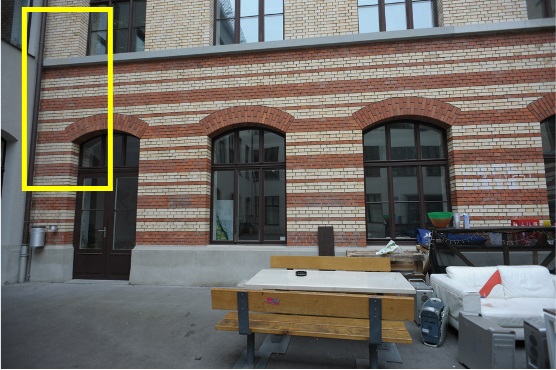}	
		}
		\caption{(b) is the reference image and the rest are source images. Obviously, (a) have more overlapping regions  with (b) while the method used by Xu may ignore it and choose (c), as there are more reconstructed sparse 3D points on the yellow regions.}
		\label{FIG:selectWithTextureLess}
	\end{figure}
	
	\begin{equation}
		\bm{\hat{X}} = (1+\epsilon)\bm{X}-\epsilon C_i 
		\label{EQ:disturb}
	\end{equation}
	\begin{equation}
		\tau = \frac {1}{\lvert A \rvert} \sum\limits _{\bm{X} \in A} \lVert   \bm{\hat{x'}}_j - \bm{x'}_j\rVert
		\label{EQ:filetrWithXu}
	\end{equation}
	
	$\bm{X}$ represents the back-projected 3D point of $\bm{x}$. First, it gives $ \bm{X}$ a disturb as  Eq. \ref{EQ:disturb} ,   $C_i$ is the camera center of reference image $i$. It computes the average projection position displacements for points in $A$ as Eq. \ref{EQ:filetrWithXu}. $ {\bm{x'}_j}$is the 2D coordinate when projecting $\bm{X}$ to  source image $j$ .   $ \bm{\hat{x'}}_j $ is the 2D coordinate  corresponding to $\bm{\hat{X}}$. The points in set $ A$ are the visible 3D points on both the reference image $i$ and the source image $j $, and these sparse 3D points are output by SFM software. The method filters source images whose  $ \tau $ is lower than $t_{\tau}$ and sorts the preserved source images with the number of points in $A$. Then it selects the top 8 images as neighboring images.  However, this neighboring image selection method may filter some suitable source images because of the lack of reconstructed points on texture-less regions and stochastic texture regions, as Fig \ref{FIG:selectWithTextureLess} shows. To overcome these weaknesses, we combine these two methods and form our method as shown in equation Eq.\ref{EQ:neighborOur}.
	\begin{equation}
		\zeta = \frac {1}{ N_A } min(b_{ij},0.1) min( \vartheta_{ij},0.05)
		\label{EQ:neighborOur}
	\end{equation}\par
	$N_A $  is the number of 3D points in $A$. We remove the source images whose $ \tau $ are lower than $t_{\tau}$ to filter source images that are too close to the reference image, and then sort the source images by the value of $ \zeta$   from small to large, then choose the smallest eight source images. Truncation equations are used here to avoid ignoring other parameters of source images whose $\vartheta$ or $b$ is too small.

	\subsection{Random Initialization}
	Good initialization can speed up the convergence of the algorithm. Therefore, we adopt the similar method as \cite{xu2020marmvs} to random initialize $d_{\bm{x}}$ using 2D feature points. It is worth noting that we only consider feature points that correspond to reconstructed sparsely  3D points produced by the SFM software. We choose the closest four feature points in four areas for each pixel, then record the minimum depth value $ d_{min} $ and maximum depth value $ d_{max}$ among four points. Then we assign a random  $d_{\bm{x}}$  within $[ d_{min} , d_{max} ]$ for  pixel $\bm{x}$. However, there may be wrong reconstructed 3D points output by the SFM software. Using wrong reconstructed points to initialize the depth may cause the result to be stuck into local optimal and cause errors. So we only consider the 3D points: (1) whose total re-project error is below 2; (2) visible to no smaller than $\lfloor v -1 \rfloor $ images, where $v $ is the average visible image number of all reconstructed 3D points in one scene; (3) the distance between its corresponding feature point to $\bm{x}$ is smaller than ${w_r}/{10} $, while  $w_r$ is the width of the reference image. To initialize the normal vector for each pixel, we adopt a similar initializing method as \cite{shen2013accurate}.  We choose  $\theta, \theta \in [0,2\pi]$ and $\phi, \phi \in [-{\pi}/{2},{\pi}/{2}]$ for each pixel and obtain the $\bm{n_x}$   as $(cos\theta sin\phi, sin\theta sin\phi, cos\phi)$.

	\begin{figure}
		\setlength{\belowcaptionskip}{-0.5cm} 
		\centering
		\includegraphics[scale=.2]{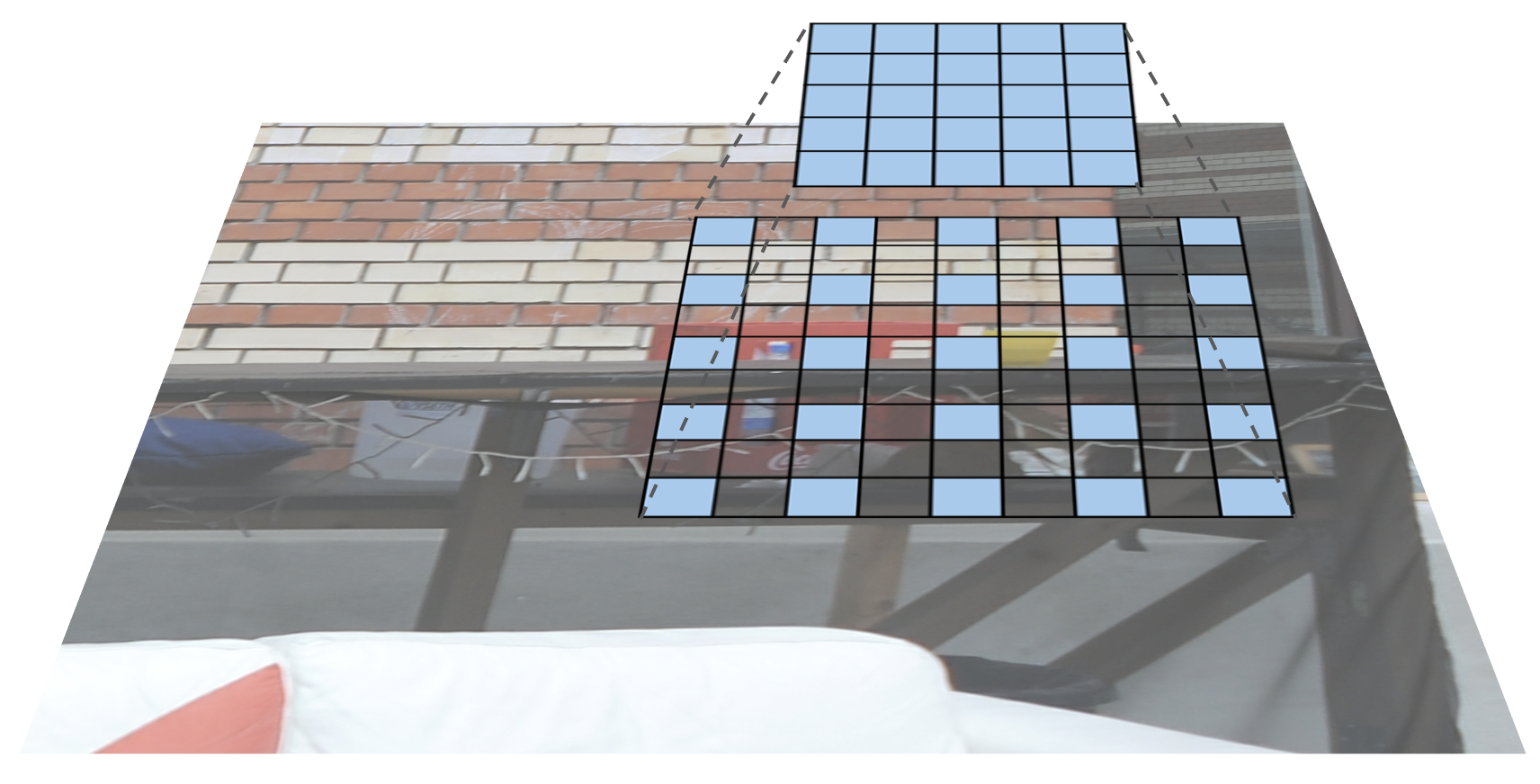}
		\caption{Schematic diagram of acceleration strategy. As the resize parameters are not necessarily integers, new pixel values will be obtained by interpolation. }
		\label{FIG:dc}
	\end{figure}
	
	\begin{figure*}
		\setlength{\belowcaptionskip}{-0.5cm} 
		\centering
		\includegraphics[scale=.3]{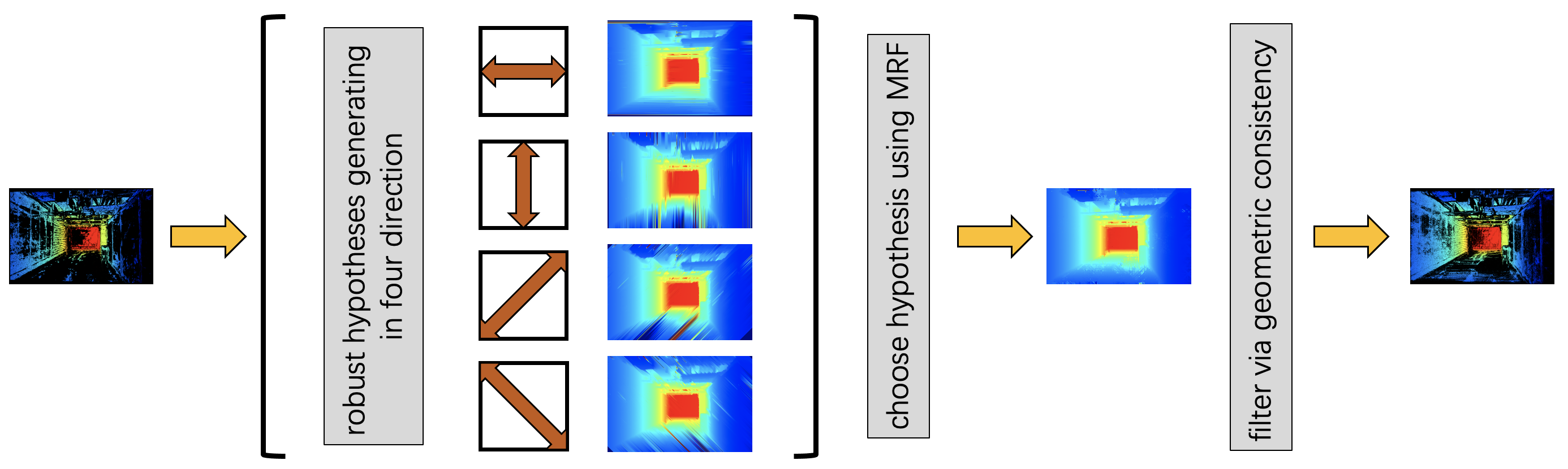}
		\caption{ The pipeline of the plane hypothesis inference procedure. First, we generate four hypotheses using the filtered depth map in four directions. Then the matching cost with each hypothesis is computed. After that, the final hypothesis is selected using MRF. Last, Geometric consistency is considered to filter outliers. }
		\label{FIG:MRFpipeline}
	\end{figure*}
	
	\subsection{Depth Map  Estimation}
	Accessing memory on the GPU is time-consuming. Sometimes, we want the matching window size larger when performing stereo matching, as using larger matching windows can make the algorithm more robust. Downsample the image is an intuitive method. However, this strategy may change the value of original pixels and may alter the structure of the original image. As we know, pixel-level differences greatly impact the reconstruction results, especially when the requirement of the result's accuracy is high.  In the experiment section, we show that downsample image is not a good choice as both the accuracy and completeness drop sharply. In order to speed up the algorithm, we borrow the idea from dilated convolution. For simplicity, we call half of the edge length of the square "window radius." We compute the scale $ s = r_{orig}/r_{now}$, $ r_{orig}$ is the original matching window radius and $r_{now}$ is the downsampled matching window radius. We compute the ZNCC values $Z(\bm{x},j)$ for each pixel with neighboring image $j $ as shown in Eq.\ref{EQ:computeDifferentWindow} and Fig.\ref{FIG:dc}.  Experiments show that our scheme is effective, and it only slightly influences the result.  \par 
	
	\begin{equation}
		Z(\bm{x},j) = \frac {\sum\limits _{\bm{\eta} \in W_{now}   }(I({\bm{\eta}})-\bar{I})(I(\bm{\eta'}_{j})-\bar{I'})}{\sqrt{\sum\limits _{\bm{\eta} \in W_{now}}{(I(\bm{\eta})-\bar{I})^2}} \sqrt{\sum\limits _{\bm{\eta} \in W_{now}}{(I(\bm{\eta'}_j)-\bar{I'})^2}}} 
		\label{EQ:computeDifferentWindow}
	\end{equation}\par
	
	$\bm{x}(x_1,x_2)$ is the 2D coordinate on images. $W_{now}$ means  new  coordinates  $\bm{\eta}(\eta_1,\eta_2)$ in the matching window as  $\eta_1 \in [x_1-sr_{now},x_1-s(r_{now}-1),\cdots,x_1+sr_{now}]$  , $\eta_2 \in [x_2-sr_{now},x_2-s(r_{now}-1),\cdots,x_2+sr_{now}]$. $\bm{\eta'_j}$ represents  the  2D coordinate on  neighboring image $j$ which can be obtained by  transforming  $\bm{\eta}$ using the homography mapping. $I(\bm{\eta})$ represents the brightness value of $\bm{\eta}$. While $\bm{\eta}$ and $\bm{\eta'}_j$ may not be integer coordinates, the $I(\bm{\eta})$ and $I(\bm{\eta'}_j)$ will be obtained by computing the bilinear interpolation.  $\bar{I}$ and $\bar{I'_j} $ are the average brightness values of $I(\bm{\eta})$ and $I(\bm{\eta'}_j)$ where $\bm{x} \in W_{now}$. 
	
	\begin{equation}
		\bm{C}(\bm{x})=\frac{\sum_{j \in N}(1-\delta(C(\bm{x},j),C_{max}))}{ \omega \sum_{j \in N}{(1-\delta(C(\bm{x},j),C_{max}))C(\bm{x},j)^{-1}} }
		\label{EQ:fcost}
	\end{equation}
	We use $\bm{C}(\bm{x})$ to represent the integrated matching cost for pixel $\bm{x}$ and $C(\bm{x},j)=1-Z(\bm{x},j)$ is the matching cost computed using  neighboring image $j$. We give restrictions on calculations. If any $\bm{\eta'}_j$  is beyond  the border of neighboring image $j$,  we set $C(\bm{x},j)$ to the maximum value $C_{max}=1-Z_{min}$. $\delta(a,b)=1$ if $a=b$ and $\delta(a,b)=0$ otherwise. Using this strategy, we can reduce the errors caused by distortion, as the distortion of the edge regions of the image will be more serious. We use the similar strategy as \cite{li2020confidence} to assign each  $C(\bm{x},j)$  a weight and the weight is $(C(\bm{x},j))^{-1}$. In summary, the final cost function is as shown in  Eq.\ref{EQ:fcost}. $N$ is the number of neighboring images.\par

	We add random disturbs to the reconstructed  result of each pixel after each spatial spreading step to help $(d_{\bm{x}},\bm{n_x})$ jumping out of the local  optimal.  We random choose  $\Delta_{d} \in [-d_{max}/4,d_{max}/4]$, $\Delta_ {\theta }\in [-\pi/2,\pi/2]$ and  $\Delta_{\phi} \in [-\pi/12,\pi/12]$ and update the $d_{old}$ and $\bm{n}_{old}$ to $d_{new}$ and $\bm{n}_{new}$. We use  4 candidates, $(d_{old},\bm{n}_{old})$, $(d_{new},\bm{n}_{old})$, $(d_{old},{\bm{n}}_{new})$, $(d_{new},{\bm{n}}_{new})$.  We update the $d$ or  $\bm{n}$ if the new candidate can generate lower matching cost value. We have also tried more candidates but it can only bring slight improvement with  computing time consuming.

	\begin{figure}
		\setlength{\belowcaptionskip}{-0.5cm} 
		\centering
		\includegraphics[scale=.23]{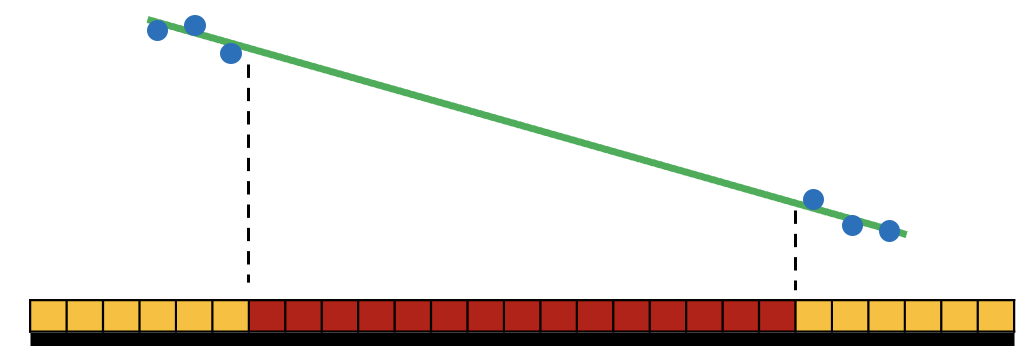}
		\caption{Orange pixels contain reconstructed depth value and red pixels not. Red pixels will be pad using the green line.}
		\label{FIG:planeFit}
	\end{figure}

	\begin{figure}
		\setlength{\belowcaptionskip}{-0.5cm} 
		\centering
		\includegraphics[scale=.25]{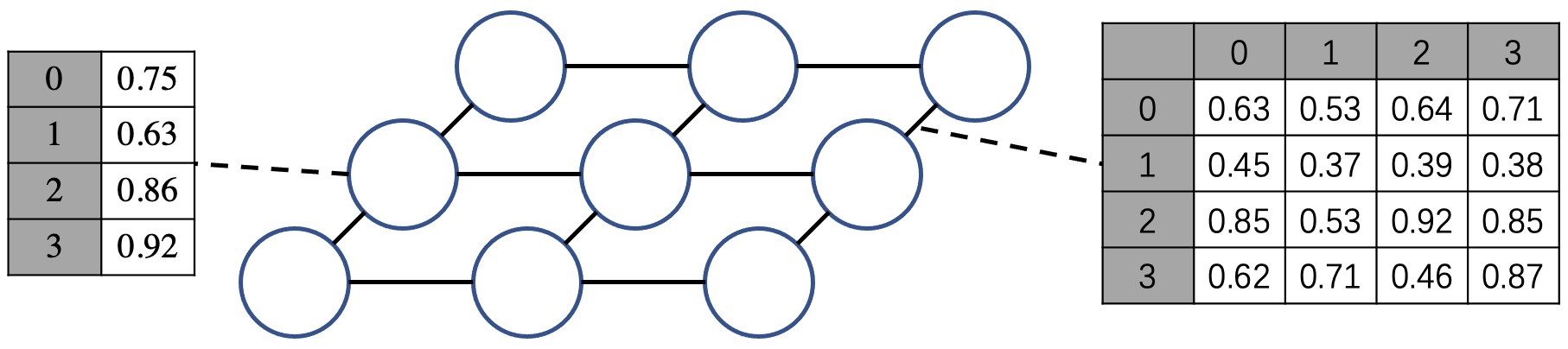}
		\caption{This is the visualization of our undirected graph model, each node represents  for each pixel, each pixel is connected with its four direct neighbors}
		\label{FIG:UGM}
	\end{figure}

	\subsection{Depth Map Filtering}
	We use the geometric consistency to filter  outliers. We adopt the similar strategy as that in \cite{schoenberger2016mvs}. $\bm{x'}_j$ is the 2D coordinate  on image $j$ corresponding to $\bm{x}$ that can be  obtained using $d_{\bm{x}}$.  $d_{\bm{x'}_j}$ is  obtained by computing bilinear interpolation and $\bm{n}_{\bm{x'}_j}$ is the same value as  $\bm{n_{[\bm{x'}_j]}}$, $[\bm{x'}_j]$ means the rounded coordinate of $\bm{x'}_j$. If the absolute difference  between $d_{\bm{x}}$ and $d_{\bm{x'}_j}$ is smaller than 0.02 , the angle between $\bm{n}_{\bm{x}}$ and $\bm{n}_{\bm{x'}_j}$ is smaller than $30^{\circ}$ and the re-project error is smaller than 1 pixel, we say  $(d_{\bm{x}},\bm{n_x})$ is consistent with $(d_{\bm{x'}_j},\bm{n}_{\bm{x'}_j})$.  If $(d_{\bm{x}},\bm{n_x})$ is  consistent with more than one $(d_{\bm{x'}_j},\bm{n}_{\bm{x'}_j})$ among all source images, we reserve $(d_{\bm{x}},\bm{n_x})$, otherwise, we discard them. \par
	
	\subsection{Plane Hypothesis Inference}
	
	It is hard to use PatchMatch based estimation methods to recover correct depth and normal vectors on texture-less plane regions. Instead of handling this issue by modifying the stereo matching step, we add a module to generate depth hypotheses on pixels that have not been reconstructed correctly. The Plane Hypothesis Inference(PHI) module mainly contains two steps: First, we generate hypotheses for each pixel; Second, we choose one hypothesis for each empty pixel as new depth. The pipeline of PHI is as shown in Fig.\ref{FIG:MRFpipeline}\par 
	To fast generate depth hypotheses, we break down the original depth map into 1D depth arrays with plenty of "holes." "Holes" mean pixels without reconstructed results.  We fill  "holes" using straight lines, the straight lines are fitted using the closest six neighboring pixels that contain reconstructed depths. We use more than two points to fit the line in order to reduce the impact of noise. As one fitted line can pad a large region of holes as Fig.\ref{FIG:planeFit} shows, our hypothesis-generating method is efficient. We generate four hypotheses for each non-reconstructed pixel in four directions: horizontal, vertical, 45 degrees below and 45 degrees above. It is shown in Fig.\ref{FIG:MRFpipeline}.\par

	\begin{equation}
		\phi_n(h(k,\bm{x})) =(C_{max}- \bm{C}(\bm{x},h(k,\bm{x})))/\kappa_1+\kappa_2
		\label{EQ:MRFU}
	\end{equation}
	
	As the accuracy of the similarity measurement methods such as ZNCC will deteriorate rapidly on texture-less regions, the MRF-based method is more suitable to be used to decide which hypothesis should be chosen. We map Markov Random Field to pairwise Undirected Graph Model(UGM) as Fig.\ref{FIG:UGM}. Each node represents a pixel on the image. For nodes, $h(k,\bm{x})$ is the $k,k\in\{1,2,3,4\}$ depth hypothesis value of $\bm{x}$, $\phi_n(h(k,\bm{x}))$ means the  value of potential function when choosing $h(k,\bm{x})$ and it is shown as Eq.\ref{EQ:MRFU}. $\kappa_1$ and $\kappa_2$ are used here to weaken the impact of the matching cost. It worth noting that the $\bm{C}(\bm{x},h(k,\bm{x}))$ represent the matching cost computed using $h(k,\bm{x})$.  Corresponding pixels on neighboring images are obtained by projecting the back-projected 3D point to neighboring images, rather than using  homography mapping. \par

	\begin{equation}
		\phi_e(h_1,h_2)=\begin{cases}
			[\kappa_3-min(1,(h_1-h_2)/h_2)]^2,& h_1>h_2\\
			[\kappa_3-min(1,(h_2-h_1)/h_1)]^2,&else
		\end{cases} 
		\label{EQ:MRFP}
	\end{equation}
	
	For edges,  the  potential function $\phi_e(h(k_1,\bm{x}_1),h(k_2,\bm{x}_2))$  is computed using the relative difference between hypothesis values of two connected nodes as Eq.\ref{EQ:MRFP}.  For  brevity, we note it as $p_e(h_1,h_2)$.  Truncation functions are used here to avoid depth values which near to the edge part over-smoothed. \par
	
	\begin{equation}
		\bm{P} = \frac {1}{Z} \prod\limits _{n } \phi_n(h(k,\bm{x}))\prod\limits _{e} \phi_e(h_1,h_2)
		\label{EQ:MRF1}
	\end{equation}

	\begin{equation}
		Z =\overbrace{\sum\limits _{k }  \cdots  \sum\limits _{k} }^{pixel \; number} \prod\limits _{n } \phi_n(h(k,\bm{x}))\prod\limits _{e} \phi_e(h_1,h_2)
		\label{EQ:MRF2}
	\end{equation}

	In pairwise UGM, the state of each node is only related to its connected nodes. The joint probability $\bm{P}$  is shown as Eq.\ref{EQ:MRF1}. Normalization constant $Z$ is simply the scalar value that forces the distribution to sum to one. In general cases, the decoding procedure of UGM is an NP-hard problem,  we use the approximate technique TRW to perform the decoding. In order to speed up the decoding process and reduce memory usage. We first downsample the hypothesis map to half of its original size, decoding UGM, and then obtaining the downsampled padded hypothesis map. Then we up-sample the padded hypothesis map and regard it as the depth reference map. Last, we choose the closest hypothesis to the hypothesis reference map for each pixel.
	
	\begin{equation}
		\bm{n}=\frac{(\bm{X}_u-\bm{X}_d) \times (\bm{X}_l-\bm{X}_r) }{||(\bm{X}_u-\bm{X}_d) \times (\bm{X}_l-\bm{X}_r)||^2}
		\label{EQ:newNorm}
	\end{equation}
	
	The normal vectors of padded pixels do not exist, and we adopt the similar normal vectors computing method as \cite{li2020confidence}. The normal vectors of these regions are computed using the four neighboring pixels of each pixel  as shown in Eq.\ref{EQ:newNorm}. $\bm{X}_u$, $\bm{X}_d$, $\bm{X}_l$, $\bm{X}_r$ are the back projected 3D points of  4-connected neighboring pixels of $\bm{x}$. After that, we again use geometric consistency to filter outliers. PHI is carried out on consumer-grade CPUs and can significantly improve the result with acceptable time consumption.\par
	
	\subsection{Depth Map Fusion}
	
	We back project each pixel on reference image $i$ to 3D space and obtain $\bm{X}_i$,  then back project the corresponding pixels whose coordinate are $[\bm{x}'_{j}]$ on all other images and obtain $\bm{X'}_j$. If the relative difference between $\bm{X}_i$ and $\bm{X'}_j$ is smaller than 0.01, we delete the corresponding pixel on image $j$ and merge $\bm{X}_i$ and $\bm{X'}_j$ by taking  the average of them. After handling all pixels on image $i$, we delete image $i$.  And then, we perform the same operation on the remaining images.
	
	\begin{algorithm}[]
		\caption{PHI-MVS pipeline} 
		\textbf{Input}: images and SFM sparse reconstruction result \\
		\textbf{Output}: merged point cloud 
		\begin{algorithmic}[1] 
			
			\STATE{/*neighboring images selection*/ }
			
			\FOR {each reference image}
			\STATE{sort  images as Eq.\ref{EQ:neighborOur}, choose top 8 as neighboring images}
			\ENDFOR
			
			\STATE{/*random initialization*/}
			\FOR {each reference image}
				\FOR {each pixel }
					\STATE{initialize depth with 4 points or random value  }
					\STATE{initialize normal vectors with  random values}
				\ENDFOR
			\ENDFOR

			\STATE{/*depth map  estimation*/ }
			\FOR {each reference image}
			
				\WHILE {iteration $\leq$ total iteration number}
					\STATE{red and black checkerboard  updating  }
					\STATE{random refinement }
				\ENDWHILE
			\ENDFOR
			\STATE{/*depth map  filtering*/ }
			\FOR {each reference image}
				\STATE{filter outliers with  geometric consistency }
			\ENDFOR
			\STATE{/*plane hypothesis inference*/ }
			\FOR {each reference image}
				\STATE{load  filtered depth map  }
				\STATE{generate depth hypotheses }
				\STATE{compute matching cost  with generated hypotheses}
				\STATE{downsample  hypotheses maps  }
				\STATE{build and decode UGM}
				\STATE{select  hypothesis using  decoding result}	
			\ENDFOR
			
			\STATE{/*depth map filtering*/ }
			\FOR {each reference image}
				\STATE{filter outliers with  geometric consistency}
			\ENDFOR
			\STATE{fusion}
			\STATE{/*fusion*/ }
			\FOR {each reference image}
				\STATE{back project pixels and merge points}
			\ENDFOR

		\end{algorithmic}
	\label{pipeline}
	\end{algorithm}
	
	\begin{figure*}
		\setlength{\belowcaptionskip}{-0.2cm} 
		\setlength{\abovecaptionskip}{-0.01cm}
		\centering
		\subfigure{
			\includegraphics[scale=.5]{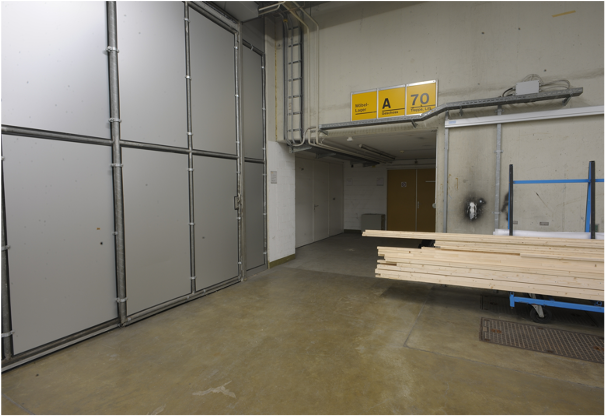}
		}
		\subfigure{
			\includegraphics[scale=.5]{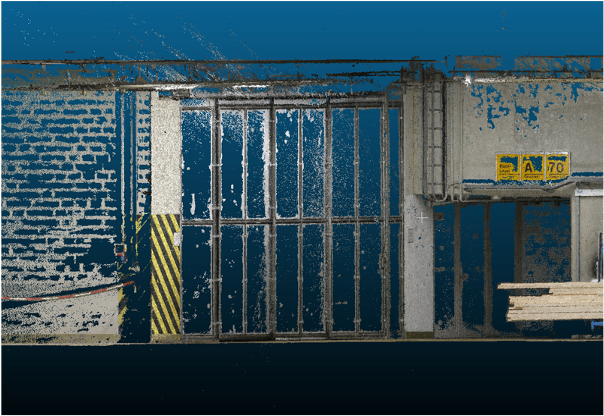}
		}
		\subfigure{
			\includegraphics[scale=.5]{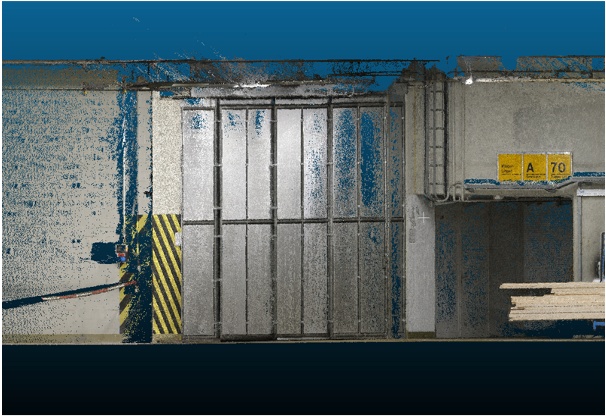}
		}
		
		\centering
		\subfigure{
			\includegraphics[scale=.5]{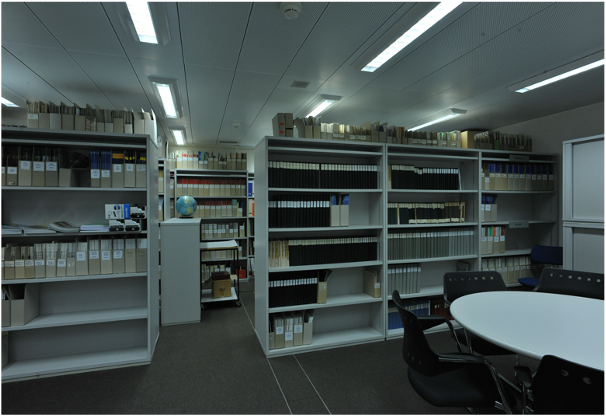}
		}
		\subfigure{
			\includegraphics[scale=.5]{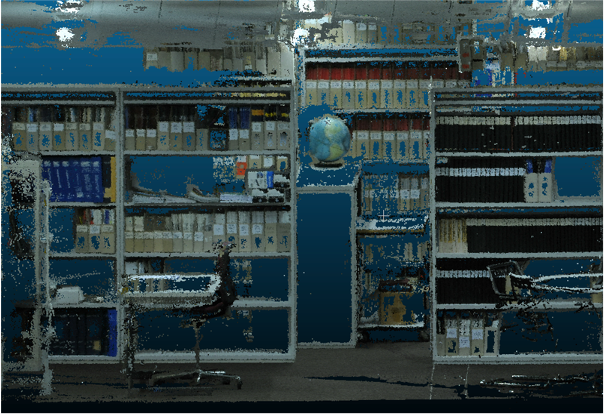}
		}
		\subfigure{
			\includegraphics[scale=.5]{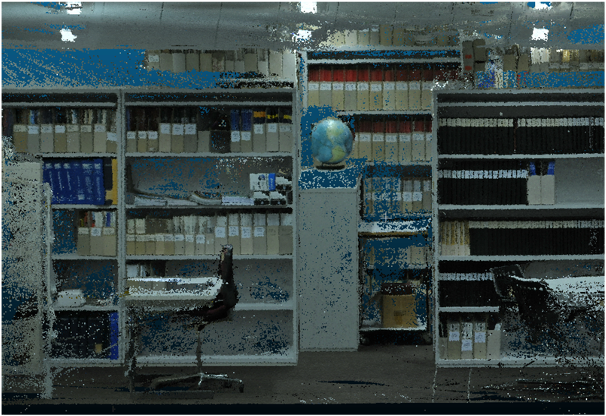}
		}
		\centering
		
		\addtocounter{subfigure}{-1}
		\addtocounter{subfigure}{-1}
		\addtocounter{subfigure}{-1}
		\addtocounter{subfigure}{-1}
		\addtocounter{subfigure}{-1}
		\addtocounter{subfigure}{-1}
		
		\subfigure[color 2D image]{
			\includegraphics[scale=.5]{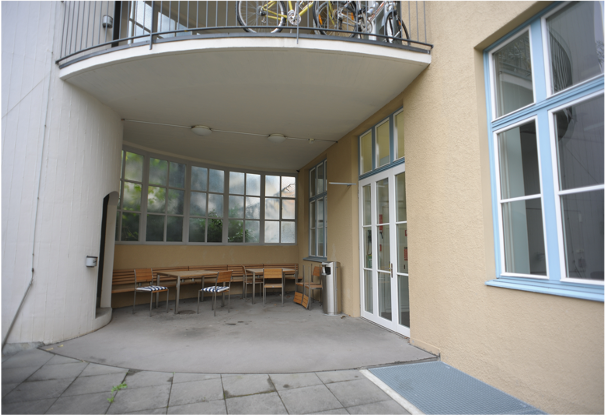}
		}
		\subfigure[point cloud without PHI]{
			\includegraphics[scale=.5]{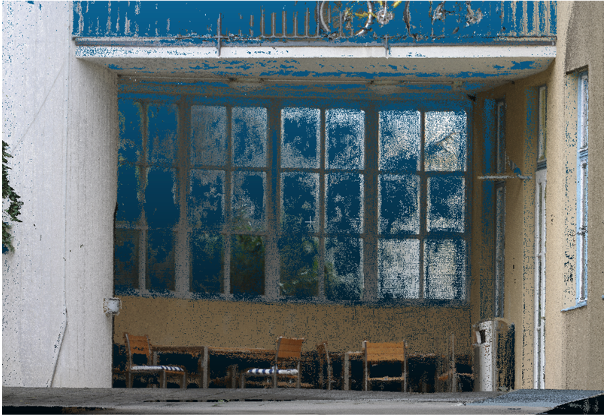}
		}
		\subfigure[point  with PHI]{
			\includegraphics[scale=.5]{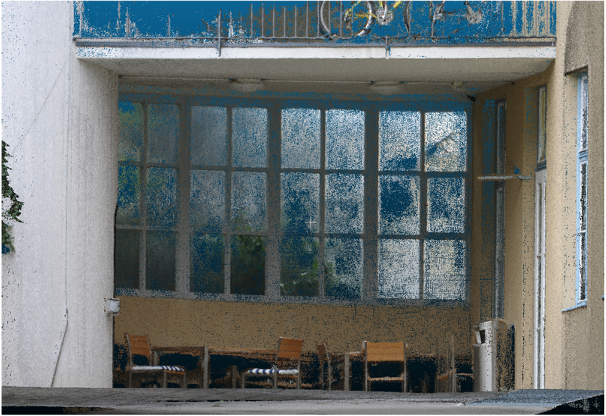}
		}
		\centering

		\caption{The first columns are input images; the second columns are the point cloud results of PHI-MVS/P; the third columns are  point cloud results of our whole pipeline. These results are reconstructed from delivery area, kicker and terrace from ETH high-resolution training set.}
		\label{FIG:trainSet}
	\end{figure*}
	
	\begin{figure*}
		\centering
		\subfigure{
			\includegraphics[scale=.095]{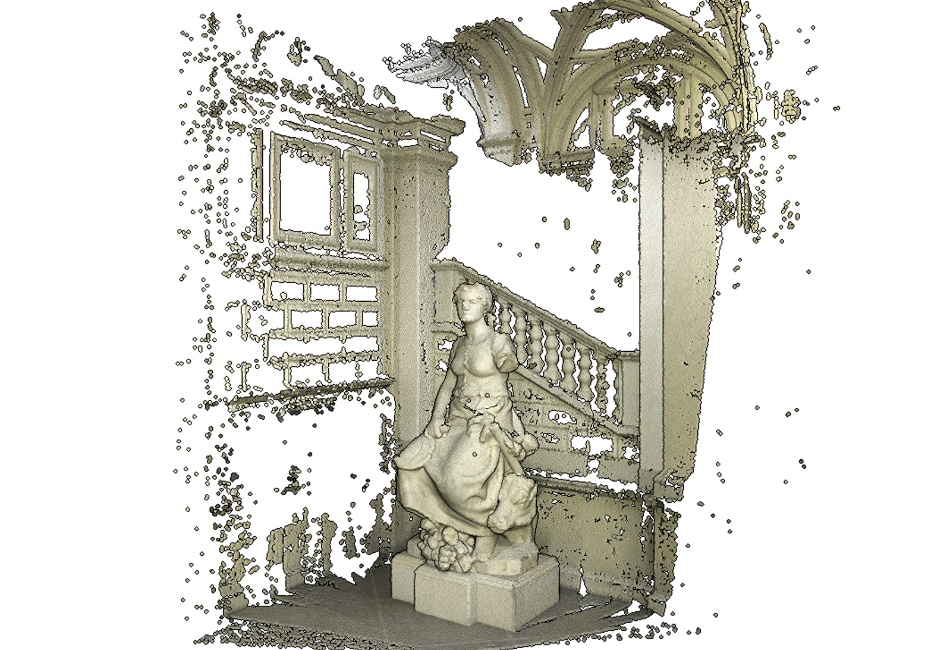}
		}
		\subfigure{
			\includegraphics[scale=.095]{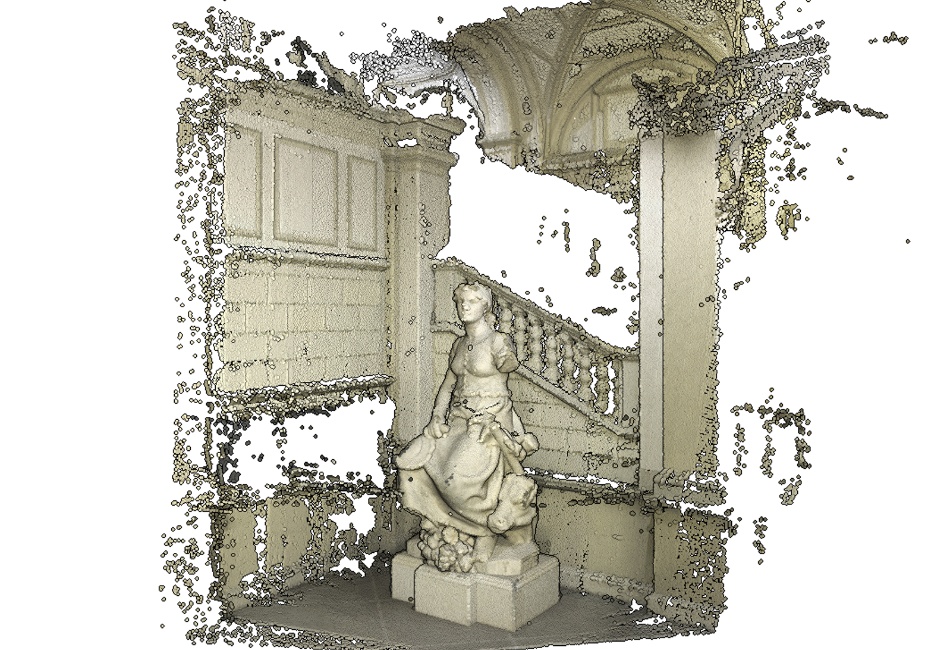}
		}
		\subfigure{
			\includegraphics[scale=.095]{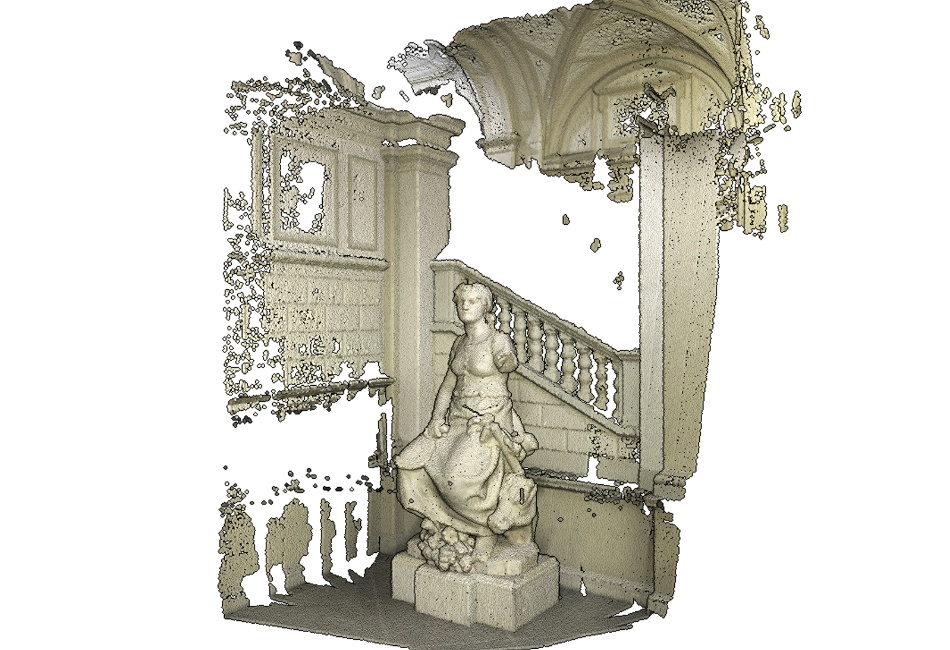}
		}
		\subfigure{
			\includegraphics[scale=.095]{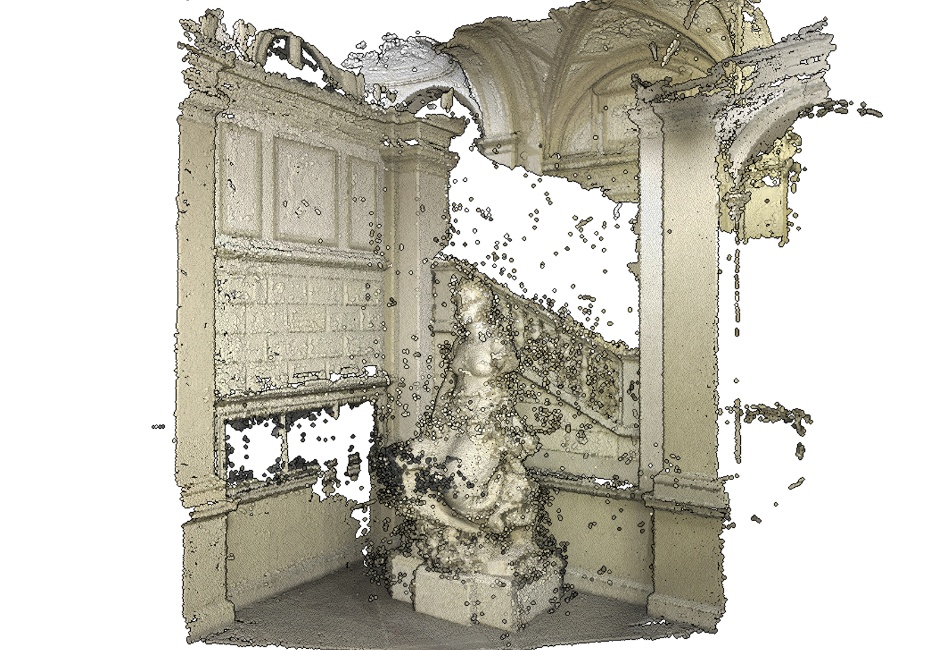}
		}
		\subfigure{
			\includegraphics[scale=.095]{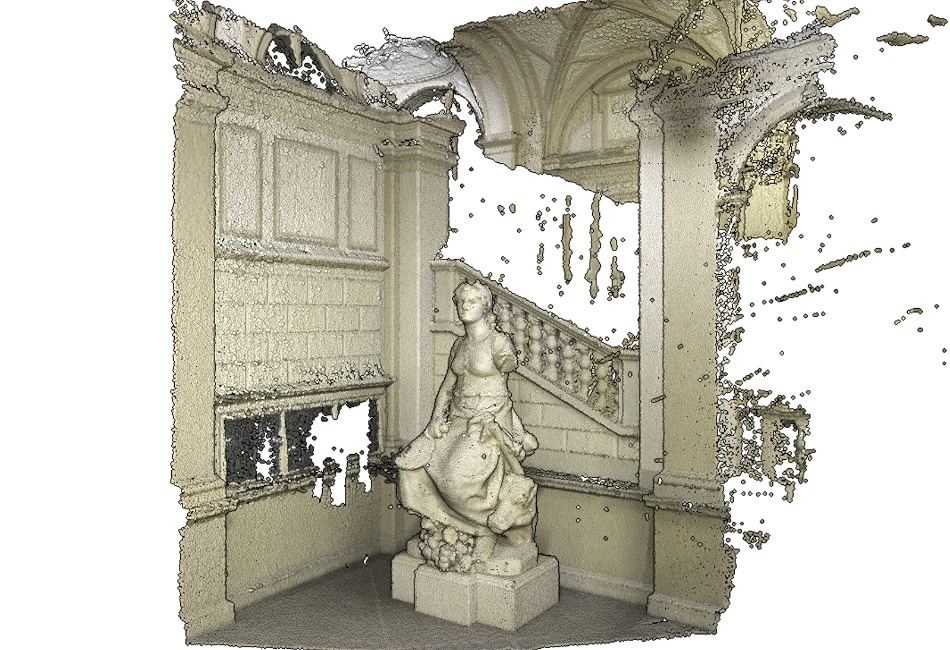}
		}
		
		\subfigure{
			\includegraphics[scale=.095]{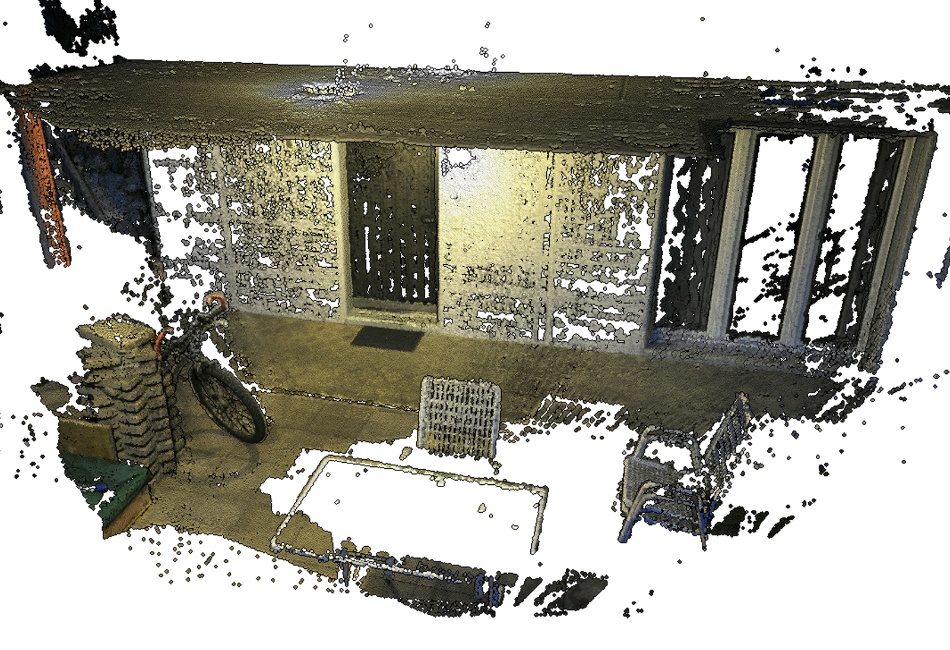}
		}
		\subfigure{
			\includegraphics[scale=.095]{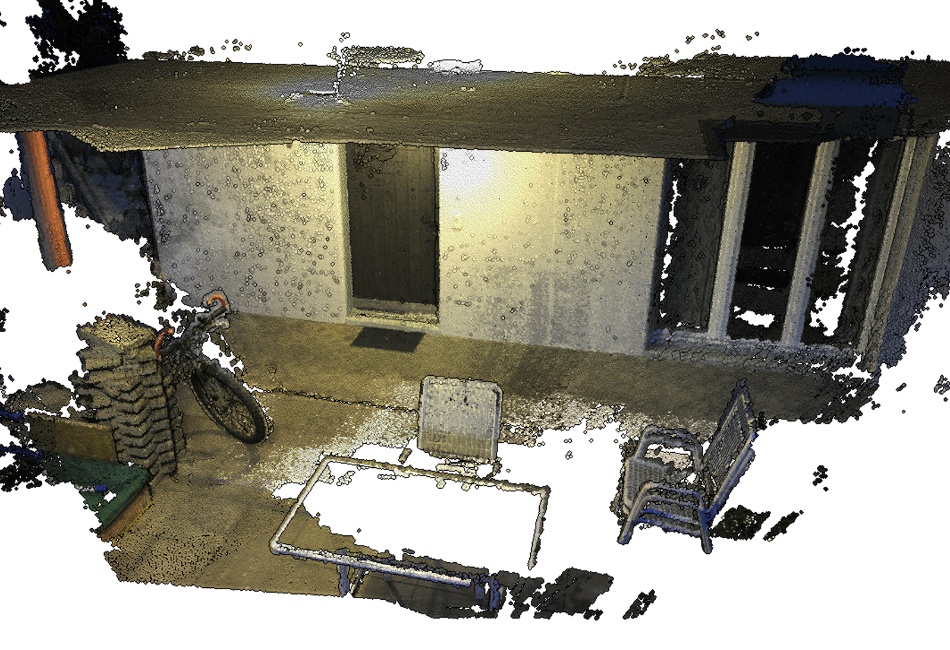}
		}
		\subfigure{
			\includegraphics[scale=.095]{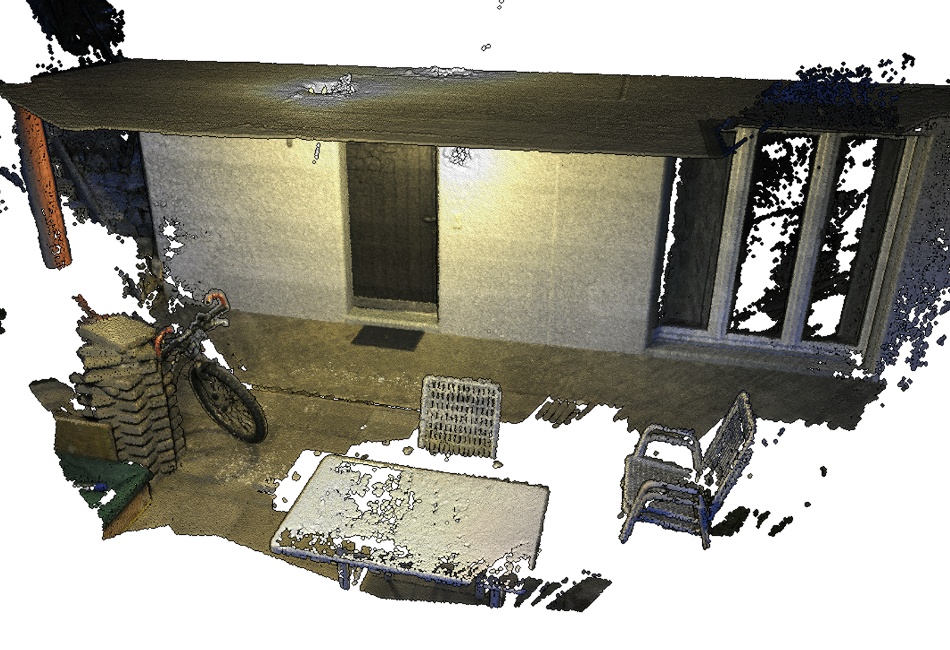}
		}
		\subfigure{
			\includegraphics[scale=.095]{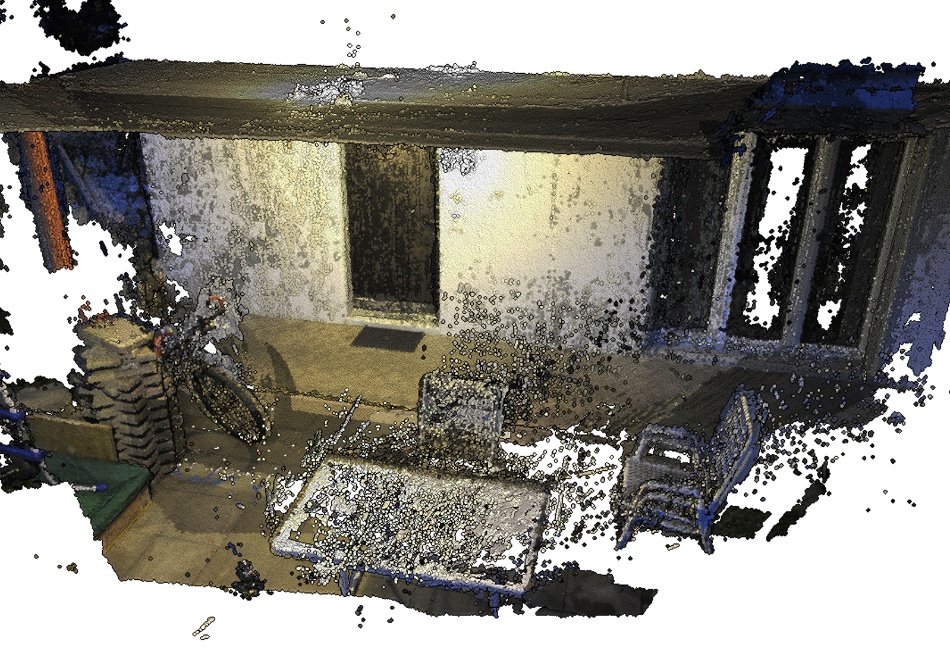}
		}
		\subfigure{
			\includegraphics[scale=.095]{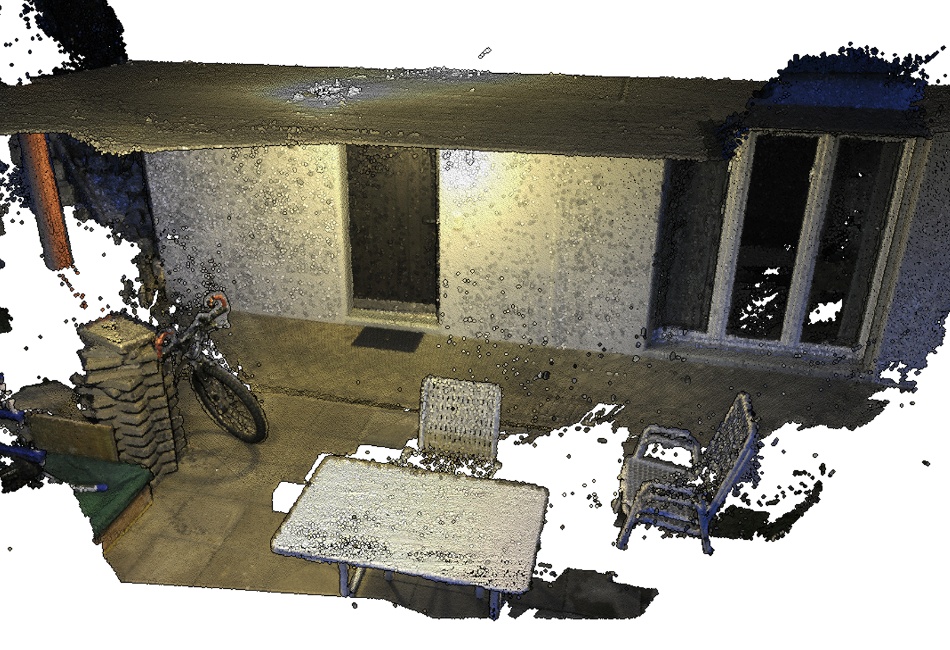}
		}
		
		\subfigure{
			\includegraphics[scale=.095]{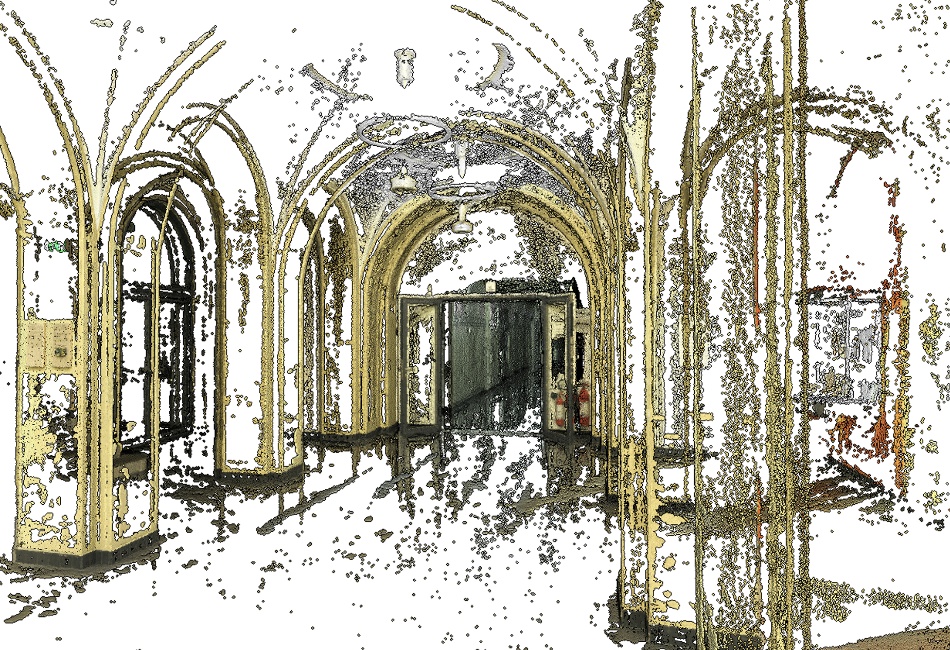}
		}
		\subfigure{
			\includegraphics[scale=.095]{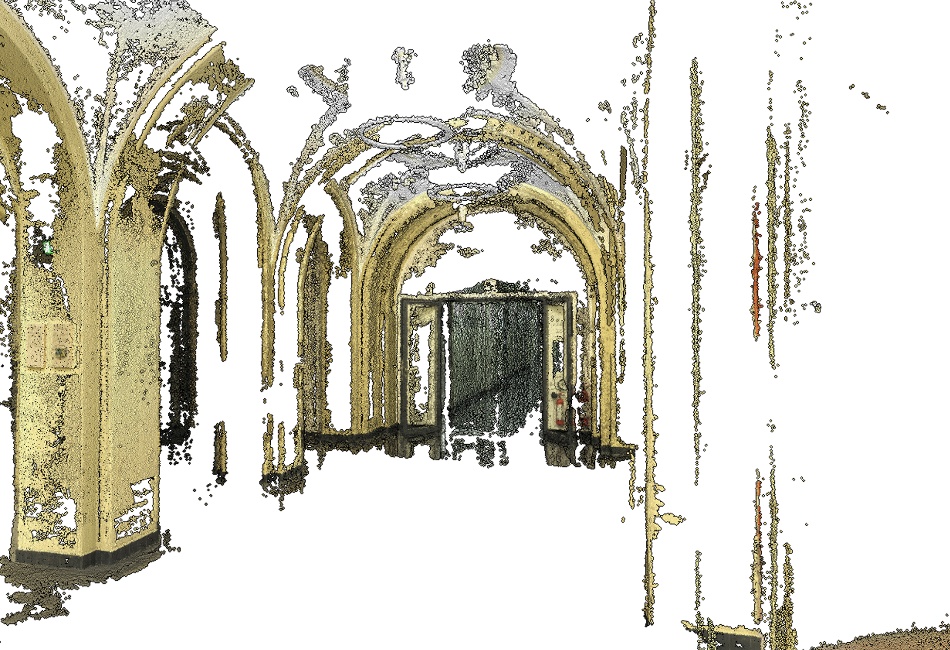}
		}
		\subfigure{
			\includegraphics[scale=.095]{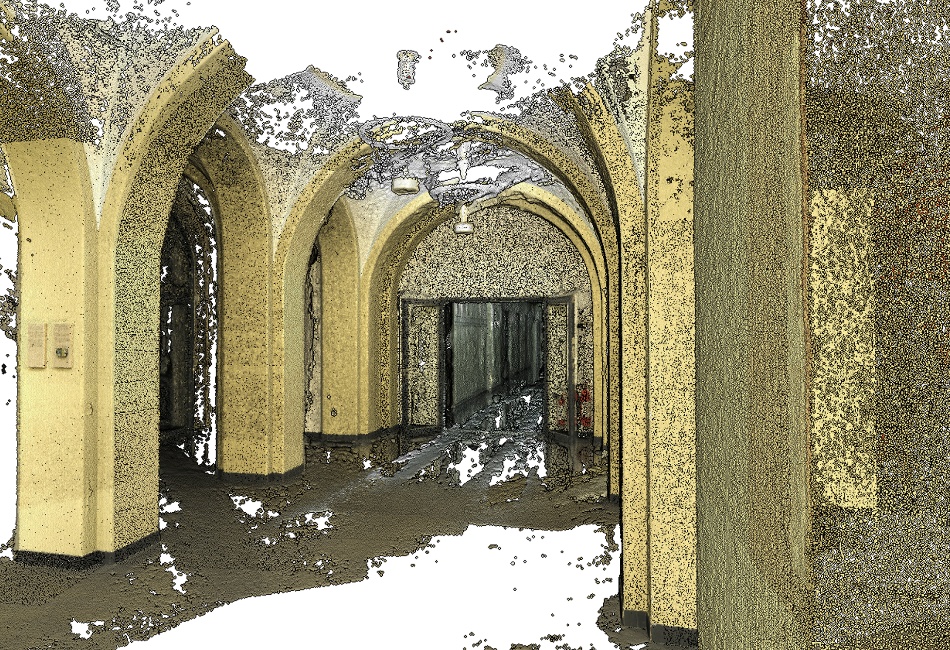}
		}
		\subfigure{
			\includegraphics[scale=.095]{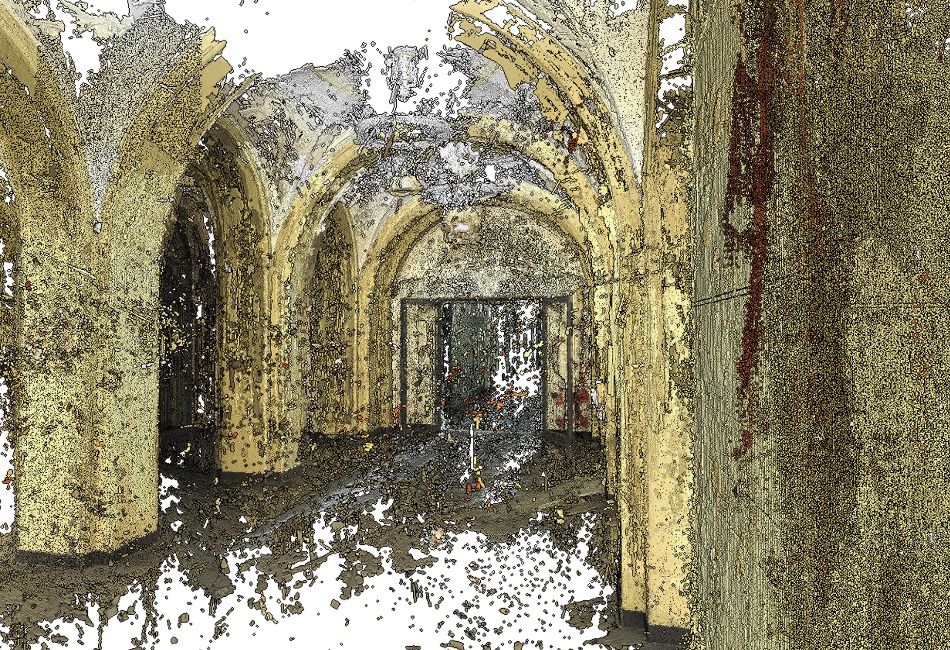}
		}
		\subfigure{
			\includegraphics[scale=.095]{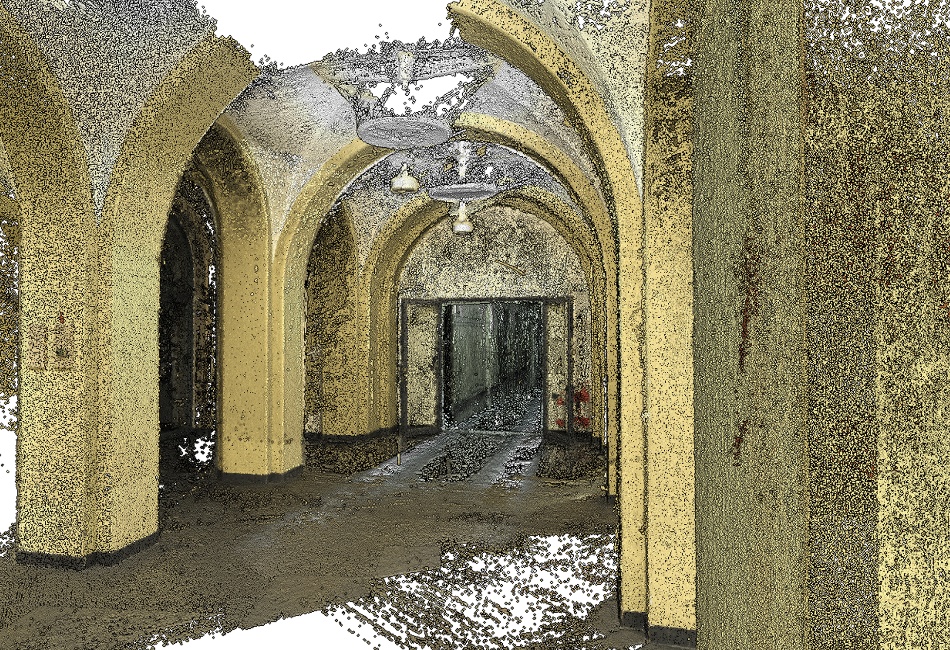}
		}
		
		\addtocounter{subfigure}{-15}
		
		\subfigure[COLMAP]{
			\includegraphics[scale=.095]{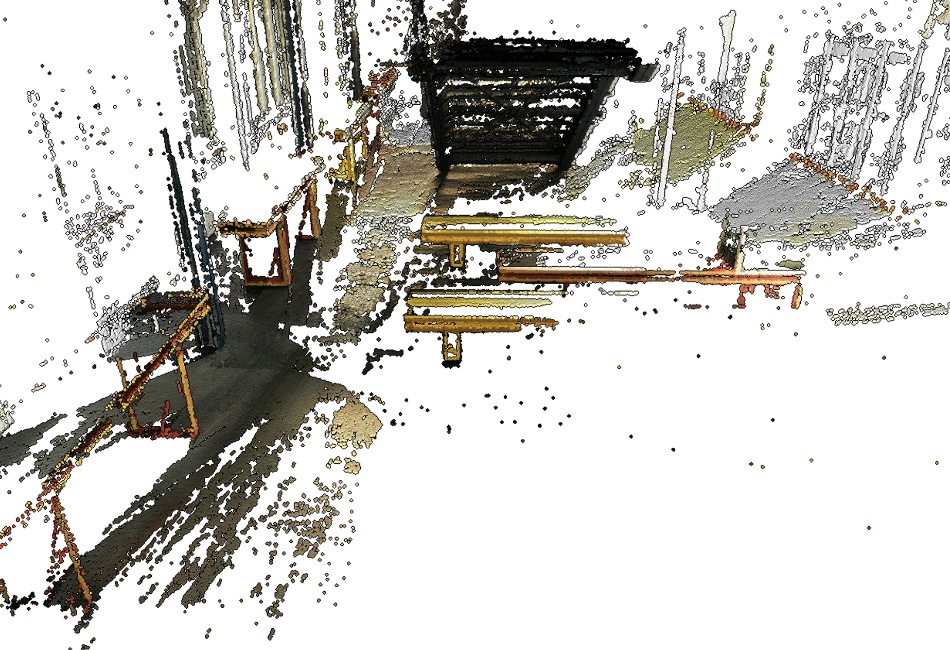}
		}
		\subfigure[TAPA-MVS]{
			\includegraphics[scale=.095]{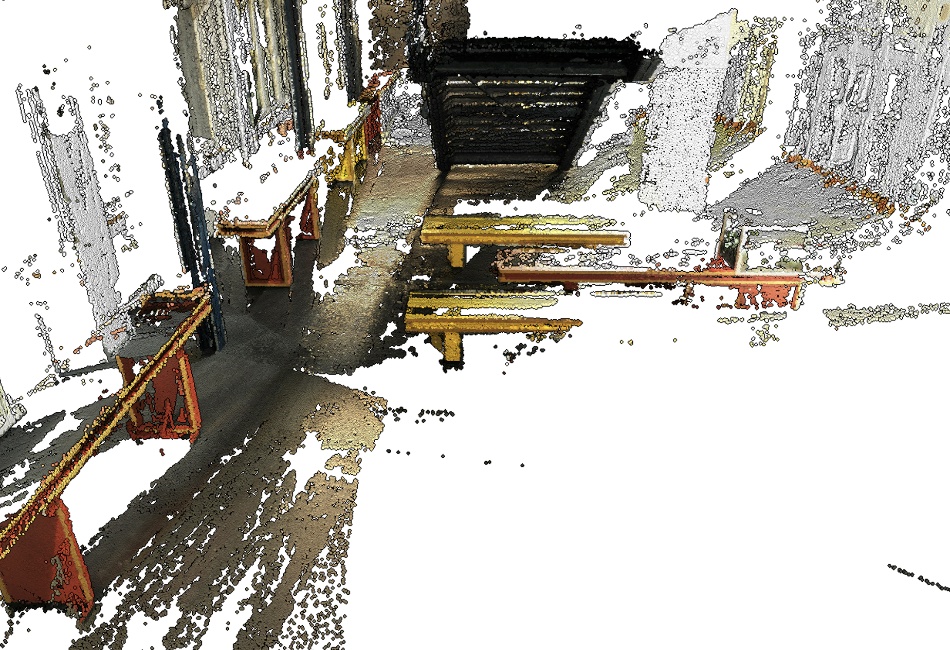}
		}
		\subfigure[ACMP]{
			\includegraphics[scale=.095]{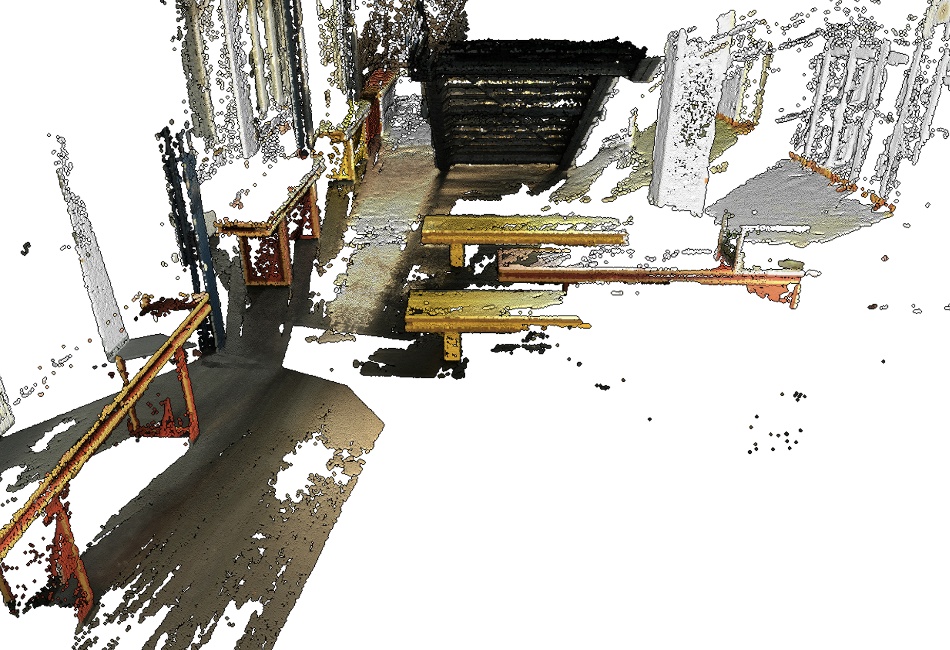}
		}
		\subfigure[MAR-MVS]{
			\includegraphics[scale=.095]{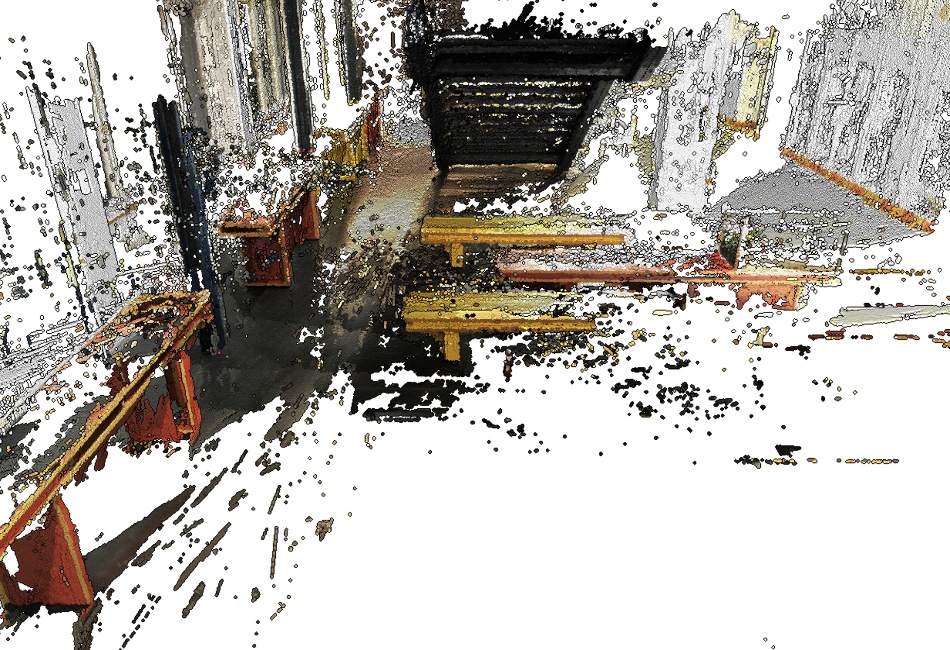}
		}
		\subfigure[PHI-MVS]{
			\includegraphics[scale=.095]{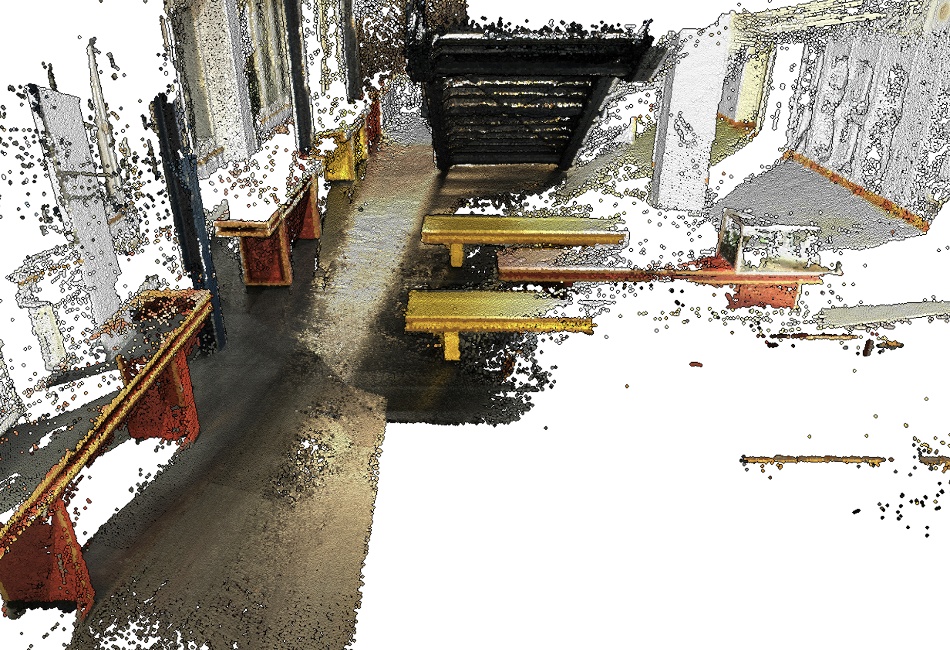}
		}
		
		\caption{Comparison to state-of-the-art algorithms on some data sets (statue, terrace, old computer and lecture) of ETH3D benchmark. These point cloud models are shown by the ETH3D  evaluation server.}
		\label{FIG:compareSOTA}
	\end{figure*}

	\begin{table*}
	
	\caption{Evaluation results on the high-resolution multi-view training datasets of ETH3D benchmark. F1 scores are evaluated at different tolerance (1cm, 2cm, 5cm and 10 cm). }
	\renewcommand{\arraystretch}{1.3}
	\centering
	
	\begin{tabular}{|p{30pt}|p{20pt}|p{20pt}|p{20pt}|p{20pt}|p{20pt}|p{20pt}|p{20pt}|p{20pt}|p{20pt}|p{20pt}|p{20pt}|p{20pt}|p{20pt}|}
		\hline
		\multirow{2}*{Data Set} & \multicolumn {4}{c|}{PHI-MVS/P} & \multicolumn {4}{c|}{PHI-MVS/M} & \multicolumn {4}{c|}{PHI-MVS} \\
		\cline{2-13}
		&1cm&2cm&5cm&10cm&1cm&2cm&5cm&10cm&1cm&2cm&5cm&10cm\\
		\hline
		courty.&65.19&85.06&93.92&96.86&66.45&86.12&94.78&97.49&\textbf{66.83}&\textbf{86.40}&\textbf{94.96}&\textbf{97.60}\\
		\hline
		delive.&70.09&85.00&94.18&97.20&73.93&89.09&96.99&\textbf{98.66}&\textbf{74.83}&\textbf{89.52}&\textbf{97.06}&98.65\\
		\hline
		electro&68.47&80.78&90.24&94.90&73.16&85.20&93.06&96.48&\textbf{73.88}&\textbf{85.52}&\textbf{93.11}&\textbf{96.49}\\
		\hline
		facade&48.25&68.66&86.82&92.46&48.86&69.23&87.65&93.08&\textbf{49.32}&\textbf{69.53}&\textbf{87.71}&\textbf{93.13}\\
		\hline
		kicker&57.97&68.63&81.33&89.83&68.68&81.74&\textbf{90.81}&\textbf{94.26}&\textbf{70.61}&\textbf{82.38}&90.71&94.07\\
		\hline
		meadow&39.28&55.46&70.11&78.26&49.45&64.34&79.67&87.16&\textbf{50.93}&\textbf{65.67}&\textbf{80.78}&\textbf{87.91}\\
		\hline
		office&49.99&61.80&75.94&85.33&67.50&80.25&92.17&96.91&\textbf{69.62}&\textbf{81.35}&\textbf{92.40}&\textbf{97.12}\\
		\hline
		pipes&59.50&68.67&82.03&91.23&73.50&81.38&90.31&96.72&\textbf{74.84}&\textbf{82.36}&\textbf{91.03}&\textbf{97.17}\\
		\hline
		playgr.&54.96&71.34&87.08&93.53&56.75&72.72&87.62&\textbf{93.92}&\textbf{57.25}&\textbf{73.07}&\textbf{87.68}&93.90\\
		\hline
		relief&73.35&84.25&91.88&95.44&75.17&86.93&94.20&\textbf{96.54}&\textbf{75.53}&\textbf{87.31}&\textbf{94.32}&96.46\\
		\hline
		relief.&71.62&83.68&92.26&95.87&73.87&86.39&94.34&\textbf{96.96}&\textbf{74.32}&\textbf{86.76}&\textbf{94.48}&96.93\\
		\hline
		terrace&75.37&86.88&94.76&98.08&77.84&89.06&96.23&\textbf{98.65}&\textbf{78.35}&\textbf{89.24}&\textbf{96.21}&98.64\\
		\hline
		terrai.&75.32&84.28&92.00&96.08&83.06&91.00&96.03&98.07&\textbf{84.17}&\textbf{91.56}&\textbf{96.16}&\textbf{98.10}\\
		\hline
		\textbf{average}&62.26&75.73&87.12&92.70&68.32&81.80&91.84&95.76&\textbf{69.27}&\textbf{82.36}&\textbf{92.05}&\textbf{95.86}\\
		\hline		
	\end{tabular}
	\label{TAB:trainMRF}
	\end{table*}

	\begin{table*}
	\caption{Evaluation results on the high-resolution multi-view test datasets of ETH3D benchmark.  F1 score, accuracy, and completeness at different tolerance (1cm, 2cm and 5cm). }\label{tbTest}
	\renewcommand{\arraystretch}{1.3}
	\centering
	\begin{tabular}{|p{65pt}|p{20pt}|p{20pt}|p{20pt}|p{20pt}|p{20pt}|p{20pt}|p{20pt}|p{20pt}|p{20pt}|p{20pt}|p{20pt}|p{20pt}|}
		\hline
		\multirow{2}*{Method} & \multicolumn {3}{c|}{1cm} & \multicolumn {3}{c|}{2cm} & \multicolumn {3}{c|}{5cm} & \multicolumn {3}{c|}{10cm}\\
		\cline{2-13}
		& F1 &  A &  C &  F1 &  A &  C &  F1 &  A & C&  F1 &  A & C\\
		\hline
		PHI-MVS&\textbf{75.8}&74.92&77.49&\textbf{86.43}&86.38&86.83&\textbf{93.89}&94.34&93.59&\textbf{96.77}&96.96&96.65\\
		\hline
		MAR-MVS\cite{xu2020marmvs}&71.51&67.72&76.76&81.84&80.24&84.18&90.30&90.32&90.63&94.22&94.21&94.43\\
		\hline
		MGMVS\cite{2020Mesh}&73.96&70.47&\textbf{78.42}&83.41&80.32&\textbf{87.11}&91.77&89.54&\textbf{94.25}&95.61&94.08&\textbf{97.24}\\
		\hline
		CLD-MVS\cite{li2020confidence}&73.33&72.88&75.60&82.31&83.18&82.73&89.47&91.82&87.97&92.98&95.47&91.06\\
		\hline
		ACMP\cite{xu2020planar}&72.30&82.64&66.24&81.51&90.54&75.58 &	89.01&95.71&84.00&92.62&97.47&88.71\\
		\hline
		ACMM\cite{xu2019multi}&70.80&82.11&64.35&80.78&90.65&74.34&89.14&96.30&83.72&92.96&98.05&88.77\\
		\hline
		PCF-MVS\cite{kuhn2019plane}&70.95&72.70&70.10&80.38&82.15&79.29&87.74&89.12&86.77&91.56&92.12&91.26\\
		\hline
		TAPA-MVS\cite{romanoni2019tapa}&66.60&75.16&61.82&79.15&85.71&74.94&88.16&92.49&85.02&92.30&94.93&90.35\\
		\hline
		COLMAP\cite{schoenberger2016mvs}&61.27&\textbf{83.75}&50.90&73.01&\textbf{91.97}&62.98&83.96&\textbf{96.75}&75.74&90.40&\textbf{98.25}&84.54\\
		\hline	
	\end{tabular}
	\label{TAB:results}
	\end{table*}

	\section{Experimental Results and Discussion}
	
	In order to show the effectiveness of our method, we have tested it on  ETH3D high-resolution benchmark\cite{schoeps2017cvpr}, whose image resolution is $6048 \times 4032$. ETH3D provides the calibration results and SFM sparse reconstruction. It uses F1 score, completeness and accuracy to measure the quality of results. ETH3D contains 13 scenes for training and 12 scenes for testing. In order to make our algorithm running within an acceptable time, we resize images to half of their original size while keeping the original aspect ratio. PHI-MVS is implemented using C++ with CUDA,  Eigen, PCL and OpenCV. For decoding UGM, we use the open-sourced library  DGM\cite{DGM}. Our algorithm is executed on a computer equipped with an Intel Core i5 9400F CPU and an NVIDIA RTX 3070 GPU. In our experiment, we verify the effectiveness of two main improvements separately. As the test set results can be upload only once a week, we only use the training set to perform the ablation study. The parameter $\{r_{now},r_{orig},\kappa_1,\kappa_2,\kappa_3,\epsilon,t_{\tau} \} $ we used are $\{5,7,4,0.5,2,0.002,0.1 \}$. The total iteration number of spatial spreading is eight. The decoding procedure of UGM is only iterated once. \par 
	
	\subsection{Plane Hypothesis Inference}  
	Our Plane Hypothesis Inference strategy can significantly improve the reconstruction of texture-less regions. Without our padding module, most of the texture-less regions could not be correctly estimated as ZNCC could not correctly measure the similarity on these regions. As shown in Fig.\ref{FIG:trainSet},  our strategy is effective, especially on some indoor scenes, which contain plenty of man-made texture planes. In order to the effectiveness of the whole PHI module, we remove the PHI module and note it as PHI-MVS/P. As we use MRF to perform the plane hypothesis selecting process, we show its effectiveness by removing the MRF, and select the hypothesis with the minimal matching cost for each pixel.  We note the pipeline without MRF as PHI-MVS/M. We test the above three pipelines on the ETH training sets, which contain 13 scenes.  As shown in Tab.\ref{TAB:trainMRF} By using PHI module, the reconstructed result can be significantly improved. Using MRF to choose hypotheses can make the result achieve a higher F1 score in low tolerance of each scene. It can make the algorithm more robust as PHI-MVS achieves the highest average F1 score under each tolerance. The total running time of the additional module is about 9s, and the most time-consuming part is the graph building step which could not be executed in parallel. The decoding procedure using TRW only costs about 2-3 seconds on the consumer-grade CPUs. The F1 score, running time all show that PHI module is efficient and effective.
	
	\begin{table}

		\setlength{\abovecaptionskip}{-0.03cm}
		\caption{Evaluation results on the high-resolution multi-view training datasets of ETH3D benchmark. F1 score and running times are evaluated at different tolerance (1cm, 2cm, 5cm and 10cm). Methods means the algorithm with different matching window radius. The first number is $r_{now}$ and the second is $r_{orig}$. }
		\label{TAB:differentwindow}
		\renewcommand{\arraystretch}{1.3}
		\centering
		
		\begin{tabular}{|p{25pt}|p{20pt}|p{20pt}|p{20pt}|p{20pt}|p{25pt}|}
			
			\hline
			\multirow{2}{*}{Method} & \multicolumn {4}{c|}{F1 Score} & \multirow{2}{*}{Time(s)} \\
			\cline{2-5}
			&1cm&2cm&5cm&10cm& \\
			\hline
			5/5&69.17&82.28&91.80&95.63&9.79\\
			\hline
			5/6&69.34&82.40&91.98&95.81&10.02\\
			\hline
			6/6&69.23&82.38&91.99&95.83&12.58\\
			\hline
			5/7&69.27&82.36&92.05&95.86&9.57\\
			\hline
			7/7&69.11&82.29&92.05&95.95&16.41\\
			\hline
			5/8&69.04&82.16&91.98&95.88&10.04\\
			\hline
			8/8&68.69&81.97&91.92&95.90&20.92\\
			\hline
			5/9&68.70&81.85&91.85&95.83&10.51\\
			\hline
			9/9&68.23&81.58&91.73&95.83&25.91\\
			\hline
			5/10&68.38&81.62&91.70&95.78&10.86\\
			\hline
			10/10&67.89&81.27&91.50&95.73&31.60\\	
			\hline
		\end{tabular}
		
	\end{table}

	\subsection{Speed up scheme}  
		We increase the matching window radius from 5 to 10, correspondingly, the window side length increases from 11 to 21. We run our whole pipeline with different $r_{now}$ and $r_{orig}$, the results of F1 scores under different tolerance and running times of depth map estimation step are shown in the Tab\ref{TAB:differentwindow}.  It can be seen that by using our speed-up scheme, the depth map estimation step can be executed faster, especially when  $r_{orig}$ is large. The result is only slightly changed comparing with that computed using the original matching window size.  We also downsample images to a quarter of their original size. Though this strategy can save computing time, the F1 score drops sharply. The F1 score drops 11.67, 6.02, 3.06, 2.00 under tolerance 1cm, 2cm, 5cm, 10cm. It can be seen that our scheme is a good choice to balance the speed and the performance.\par

	\subsection{Overall Evaluation}
	The evaluation of our algorithm's whole pipeline is performed on the test set of the ETH3D high-resolution multi-view dataset. Our result has achieved state-of-the-art results among the published algorithms. Besides, our algorithm is efficient and can be executed within an acceptable time. The comparison with other algorithms is shown in Fig.\ref{FIG:compareSOTA} and Table\ref{TAB:results}. It can be seen that our new pipeline can obtain a better trade-off between accuracy and completeness.
	
	\section{Conclusion}
	In this paper, we  present a novel MVS pipeline which is called Plane Hypothesis Inference Multi-View Stereo(PHI-MVS). The additional PHI module can significantly improve the reconstruction of texture-less regions in two steps. First, it generates multiple plane hypotheses from initial depth maps using a robust padding strategy, and then chooses depth hypotheses for each pixel using  Markov Random Field. Besides, a speed-up scheme similar to dilated convolution is used to speed up our algorithm. Experimental results on the ETH3D benchmark show that PHI-MVS is efficient and effective. In future works, we plan to improve the accuracy of the reconstructed results using local texture information.

	\bibliographystyle{IEEEtran}

	\bibliography{myReference}

\end{document}